\documentclass[10pt,journal,compsoc]{IEEEtran}
\usepackage{cite}
\usepackage{amsmath}
\usepackage{amssymb}
\usepackage{pifont}
\newcommand{\cmark}{\ding{51}}%
\newcommand{\xmark}{\ding{55}}%
\usepackage{amsmath,amssymb,amsfonts}
\usepackage{algorithmic}
\usepackage[colorlinks=true,linkcolor=blue,citecolor=blue]{hyperref}
\usepackage{graphicx}
\usepackage{textcomp}
\usepackage{color}
\definecolor{azure}{rgb}{0.0, 0.5, 1.0}
\usepackage{booktabs}
\usepackage[flushleft]{threeparttable}
\usepackage{marginnote}
\usepackage{booktabs, makecell, multirow, tabularx}

\ifCLASSOPTIONcompsoc
\else
\fi
\ifCLASSINFOpdf
\else
\fi
\begin{document}
%
\title{Graph Neural Networks in Network Neuroscience}
%
%
%
%
\author{Alaa Bessadok, Mohamed Ali Mahjoub, and Islem Rekik, \IEEEmembership{Member, IEEE}
\thanks{A. Bessadok and I. Rekik are affiliated with BASIRA Lab, Faculty of Computer and Informatics, Istanbul Technical University, Istanbul, Turkey. Corresponding author: Islem Rekik, \url{www.basira-lab.com}}
\thanks{I. Rekik is co-affiliated with the School of Science and Engineering, Computing, University of Dundee, UK. Email: irekik@itu.edu.tr}
\thanks{A. Bessadok and MA. Mahjoub are affiliated with LATIS Lab, ISITCOM, ENISo, University of Sousse, Sousse, Tunisia. Email: alaa.bessadok@gmail.com mohamedali.mahjoub@eniso.rnu.tn}
}

\author{Alaa~Bessadok,
        Mohamed~Ali~Mahjoub,
        and~Islem~Rekik,~\IEEEmembership{Member,~IEEE,}
\IEEEcompsocitemizethanks{\IEEEcompsocthanksitem A. Bessadok and I. Rekik are affiliated with BASIRA Lab, Faculty of Computer and Informatics, Istanbul Technical University, Istanbul, Turkey. Corresponding author: Islem Rekik, \url{www.basira-lab.com}
\IEEEcompsocthanksitem I. Rekik is co-affiliated with the School of Science and Engineering, Computing, University of Dundee, UK. \protect\\
E-mail: irekik@itu.edu.tr
\IEEEcompsocthanksitem A. Bessadok and MA. Mahjoub are affiliated with LATIS Lab, ISITCOM, ENISo, University of Sousse, Sousse, Tunisia.\protect\\E-mail: alaa.bessadok@gmail.com mohamedali.mahjoub@eniso.rnu.tn}

}

%
%

\markboth{Journal of \LaTeX\ Class Files,~Vol.~14, No.~8, August~2015}%
{Shell \MakeLowercase{\textit{et al.}}: Bare Demo of IEEEtran.cls for Computer Society Journals}
%



\IEEEtitleabstractindextext{%
\begin{abstract}
Noninvasive medical neuroimaging has yielded many discoveries about the brain connectivity. Several substantial techniques mapping morphological, structural and functional brain connectivities were developed to create a comprehensive road map of neuronal activities in the human brain --namely brain graph. Relying on its non-Euclidean data type, graph neural network (GNN) provides a clever way of learning the deep graph structure and it is rapidly becoming the state-of-the-art leading to enhanced performance in various network neuroscience tasks. Here we review current GNN-based methods, highlighting the ways that they have been used in several applications related to brain graphs such as missing brain graph synthesis and disease classification. We conclude by charting a path toward a better application of GNN models in network neuroscience field for neurological disorder diagnosis and population graph integration. The list of papers cited in our work is available at \url{https://github.com/basiralab/GNNs-in-Network-Neuroscience}.
\end{abstract}

}

\maketitle

\IEEEdisplaynontitleabstractindextext

%
\IEEEpeerreviewmaketitle

\IEEEraisesectionheading{\section{Introduction}\label{sec:introduction}}

%
%
%
%

\IEEEPARstart{U}{nderstanding} how the underpinning connections in the biological neural system account for human brain functions has been a preoccupation of researchers for several years \cite{Bassett2017,Basset2008,Lydon-Staley2018}. To accomplish this aim, network neuroscience field has emerged with the aim of mapping the endogenous activity generated by neurons. Particularly, neuroscientists modeled the brain as a graph where nodes represent the anatomical brain regions connected by edges referring to morphological, functional, or structural connectivities relating pairwise nodes \cite{Fleischer2019,Fornito2016,vandenHeuvel2019}. Conventionally, three types of brain graphs can be defined from different medical modalities: morphological, structural, and functional brain graphs, which are derived from structural T1-weighted, diffusion-weighted and resting-state functional MRI, respectively \cite{Mahjoub2018,Booth2015,Stam2004,Hagmann2008,Logothetis2008}. Such graph representation of the brain gives rise to the understanding of the life span of brain wiring across three axes. The first one represents the \emph{time} axis where we can track the brain connectivity changes over time from birth to age or foresee a transition from a healthy to a disordered state. By deepening our understanding of the relationship between the brain connectivities at a first timepoint and its follow-up acquisitions, we may produce novel diagnostic tools for better identifying neurological disorders at a very early stage (e.g., Alzheimer's disease and autism) \cite{time1,time2}. The second axis refers to the \emph{resolution} of the brain graph which yields to discovering new inter-regional connections that hold the brain. Specifically, a brain can be modeled by a low-resolution or super-resolution connectome where the former one has fewer nodes (i.e., region of interest (ROIs)) and edges than the latter. Mainly, each connectome scale represents its own specific window into the topographic organization of the brain \cite{Fornito2015}. Thus, this diversity in resolution will certainly increase the performance of early disease diagnosis \cite{MhiriSuper} as it helps better capture the multi-level nested complexity of the brain as a connectome. Ultimately, knowing that neurological disorders affect the brain in different ways, one may boost the diagnosis by leveraging the complementary information present in multiple modalities such as functional and morphological connectivities \cite{Mahjoub2018,Soussia2018,Soussia2018b}. Therefore, the third axis refers to the \emph{domain} in which the brain data was collected, which is commonly referred to in network neuroscience as the `neuroimaging modality' (e.g., functional MRI or diffusion MRI) utilized to generate the brain connectome type (e.g., functional or structural). Such cross-domain multimodal representation of the brain connectivity provides invaluable complementary information for brain mapping in health and disease.

\begin{figure}
\centerline{\includegraphics[width=\columnwidth]{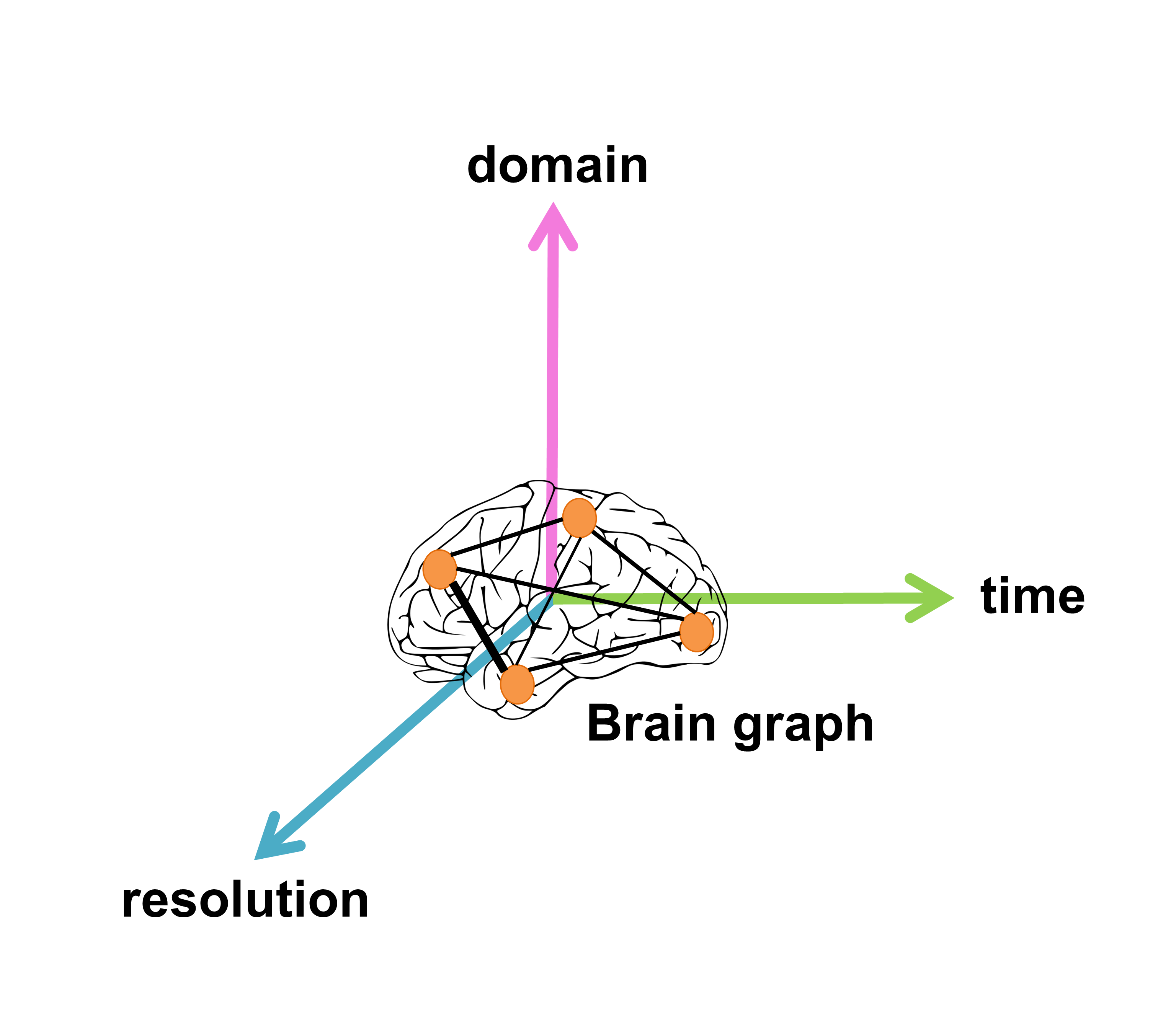}}\caption{Three axes to generate brain graphs. From a single brain graph, existing works can generate different types of brain connectivities (domain axis), produce higher resolution graphs (resolution axis) and predict graph topological changes over time (time axis). }
\end{figure}

Ideally, one would have a connectomic dataset that spans all these dimensions to benefit from the complementary between different brain network representations. Such multimodal dataset will provide rich information that helps generate a holistic connectional roadmap encoding different facets of the brain thereby understanding typical and atypical brain connectivities \cite{Lydon-Staley2018,vandenHeuvel2019}. Several connectomic datasets have been proliferated such as the Human Connectome Project (HCP) \cite{hcp}, the Baby Connectome Project (BCP) \cite{babyconnectomeUNC} and the Connectome Related to Human Disease (CRHD) \cite{crhd}. In these datasets, connectivity matrices were estimated using different tools. For instance, functional connectivity matrices were generated using CONN toolbox or groupwise whole-brain parcellation approaches \cite{whitfield2012conn,shen2013groupwise}. On the other hand, structural and morphological connectivity matrices were measured using FSL toolbox and Desikan-Killiany atlas via FreeSurfer software \cite{jenkinson2012fsl,desikan2006automated}, respectively. Although existing connectomic datasets were used to evaluate various GNN-based frameworks referred in our review, they often have missing observations due to various reasons such as high-cost of clinical scans and time-consuming preprocessing step of neuroimaging data.

To overcome this obstacle, various graph generative models emerged \cite{Minghui2018,Ghribi,Cengiz2019,MhiriSuper,Ezzine} based on different machine learning (ML) architectures. Such models aim to generate a holistic mapping of the brain from minimal resources where the prediction of multimodal \cite{Minghui2018}, high-resolution \cite{Cengiz2019,MhiriSuper} and temporal evolution \cite{Ghribi, Ezzine} of brain networks is achieved from single-domain, single-time and single-scale source brain graph, respectively. Leveraging such brain graph generative models in a real clinical setting will be beneficial in several aspects: for a disordered subject having only a morphological brain graph derived from T1-weighted MRI, one can chart out all the missing modalities (diffusion and functional) over time and at various resolutions. This will reduce the scanning time as well as the time-consuming neuroimage processing pipelines. Basically, from one single baseline brain graph, we can predict all its variations across different domains, resolutions and timepoints. Another line of ML-based generative models have been implemented via graph integration \cite{Mhiri2020,Saglam2020,Rekik2017,Dhifallah2019}. The integration aims to generate a representative template of a population of brain muligraphs with a shared neurological state (e.g., autistic). This generated connectional brain template (CBT) encodes a holistic mapping of shared traits within a population of brain multigraphs.

Another set of ML methods have been developed for learning the embedding of brain graphs into different spaces such as geometric and hyperbolic spaces, thereby enabling a better visualization of the complex topology of brain connectivity. {For instance, \cite{ye2015intrinsic} proposed BRAINtrinsic which represents the first dimensionality reduction study without inferring anything substantive about the intrinsic geometry of brain connectivity.} Owing to previous works which demonstrated that linear dimensionality reduction techniques such as principal component analysis (PCA) have been used to explore biomarkers related to various diseases such as breast cancer, \cite{ye2015intrinsic} proposed a nonlinear dimensionality reduction technique which embeds the brain by considering the topology depicted in the connectivity matrix rather than the anatomic distances. While novel, this brain graph embedding technique is minimally related to neuroanatomy. {Later on, \cite{cacciola2017coalescent} proposed an alternative to brain graph embedding in the geometric space by learning the coalescent embedding of brain connectivities in the hyperbolic space. The proposed method successfully segregated the brain graph into spatially distinct subgraphs representing the brain lobes. Mainly, the proposed work represents the first network geometry markers for brain diseases and the coalescent embedding allowed the detection of geometrical pathological changes in the connectomes of Parkinson’s Disease patients with respect to control subjects.} \cite{allard2020navigable, zheng2020geometric} have further demonstrated the hyperbolic geometry has led to valuable mapping of brain graphs, where an essential finding is that the closer coordinates in hyperbolic space denoting nodes in the graph the more likely to be connected. More recently, \cite{whihyperbolic} proposed a framework to discover the best geometry space for functional brain graphs where brain graphs were embedded onto 2-D hyperbolic discs. Interestingly, evaluating this method on disordered subjects demonstrated the existing of abnormal pathways by comparing internodal hyperbolic distances. The importance of such brain connectivity spatial embedding methods has been demonstrated in several applications such as graph navigation \cite{seguin2018navigation, muscoloni2019navigability} where the goal is modeling communication strategy within the graph that propagates signals based on the distance between nodes. 

A separate line of work leveraged deep learning (DL) models such as convolutional neural network (CNN) for learning brain graphs \cite{Chen2020}. An artificial neural network (ANN) is generally represented as a graph of connections between neurons. Based on this definition a recent study \cite{you2020graph} was conducted to understand the relationship between the graph structure of the neural network and its predictive performance. To achieve this, the authors of the paper proposed to create a relational graph which represents the neural network: the link between nodes does not represent the data flow as in the computational graph of ANN but a message exchange in the relational graph. Interestingly, they discovered that the relational graph that leads to improvement of the predictive performance of ANN were highly similar to brain connectomes. While effective when compared to traditional machine learning methods, still DL methods do not generalize well to \emph{non-Euclidean} data types (e.g., graphs). More specifically, directly applying DL methods to graphs overlooks the relationships between nodes and their local connectedness patterns. This causes an important loss of topological properties inherently encoded in a graph representation.

Recently, graph neural networks (GNNs) have been proposed to tackle this issue. GNNs are the core of a nascent field dealing with various graph-related tasks such as graph classification and graph representation learning. The main advantage of this domain is that it preserves graph topological properties while learning to perform a given task \cite{Zhou2018,Wu2020,Zhang2020}. Such learning frameworks encapsulate both graph representation learning via embedding and scoring prediction via different operations such as pooling. The first GNN was designed for node classification where each node in the graph is associated with a label \cite{Scarselli2008}. It mainly performs a propagation rule encapsulating node and edge features to generate prediction scores. Notably, the brain graph is a natural fit for GNN models thus there is a need to outline the reason of GNNs being worth investigating in network neuroscience and highlight a selection of current and future works. While several review papers already exist, they are all different from our review. For instance, \cite{Zhou2018,Wu2020,Zhang2020,Bronstein2017,Liu2020} and \cite{Bassett2017,Basset2008,Lydon-Staley2018,Fleischer2019,Fornito2016,vandenHeuvel2019,Fornito2015,Soussia2018,Brown2016} do not discuss specific GNN-based methods for solving neuroscience problems, but instead act as a reference for a specific topic (i.e., GNN or neuroscience). Therefore, it is necessary to provide a high-quality review that analyzes the trends and highlights the future directions for the applications of GNNs to the field of connectomics, which can generalize to the broader field of ``omics" (e.g., genomics) \cite{genomic1,genomic2}. We particularly conduct in this survey a comprehensive review of graph learning tasks such as graph prediction, classification and integration with application to network neuroscience where the main data structure is brain graph.

The rest of this review is organized as follows. In Section 2, we first give an overview of existing GNN models. Next, we propose a taxonomy of existing models and discuss the limitations of each category. Finally, in Section 3 we propose several future research directions and we conclude by summarizing our findings from this study.

\begin{figure*}
\centerline{\includegraphics[width=2\columnwidth]{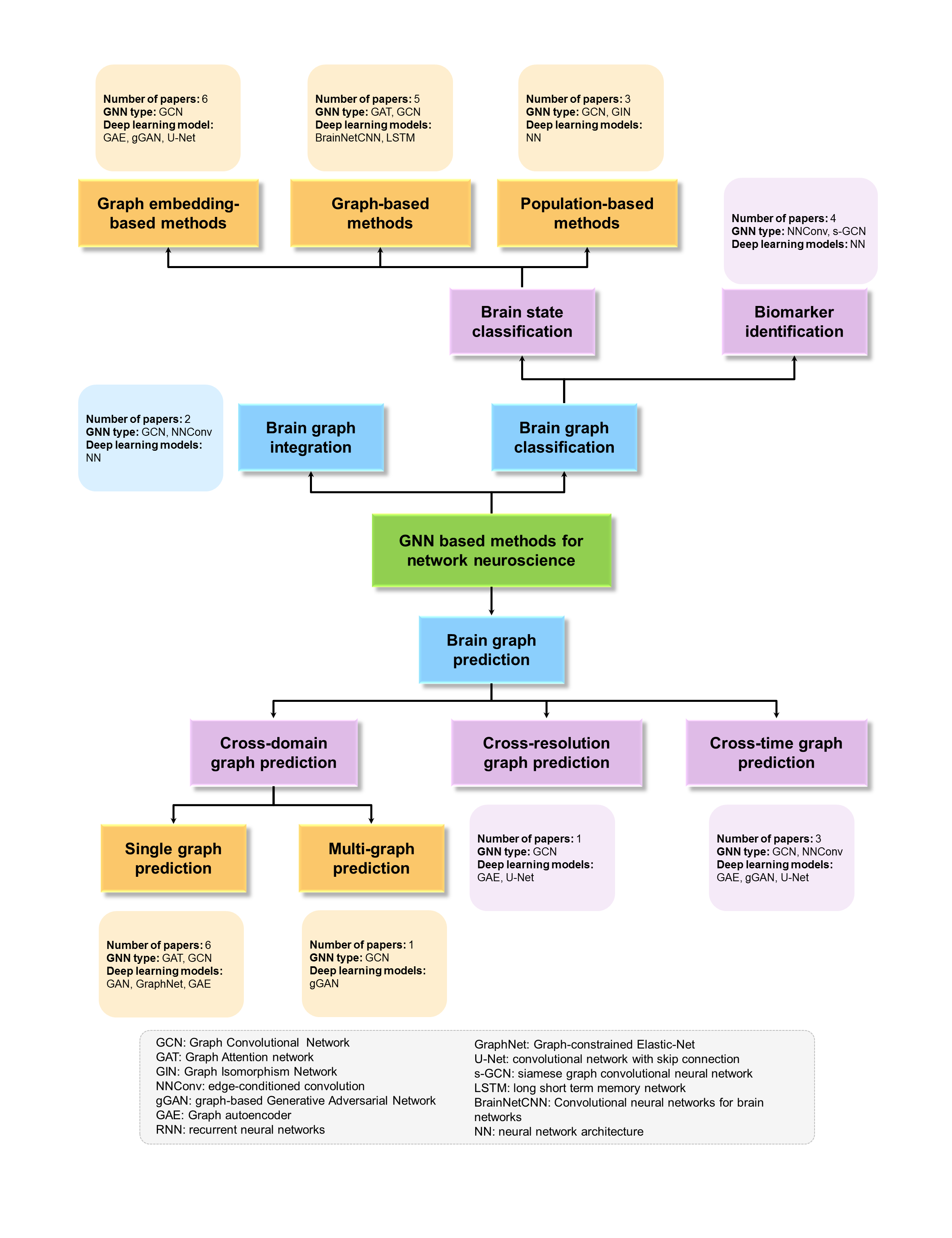}}
\caption{Categorization landscape of GNN-based methods for network neuroscience. Existing methods fall into three broad categories: brain graph prediction (blue box down), disease classification (blue box top-right) and brain graph integration (blue box middle-left) approaches.}
\label{diagram}
\end{figure*}

\section{WHAT DO GRAPH NEURAL NETWORKS OFFER TO NETWORK NEUROSCIENCE?}

\subsection{GNN overview}

GNNs are among the hottest topics nowadays thanks to their benefits in learning on a wide range of graphs, both directed and undirected. GNNs have emerged as subfield of the broader field of ``geometric deep learning" on non-Euclidean data. Broadly, GNNs can be categorized into three big classes \cite{Bronstein2017}, either by aggregating the features of neighborhood nodes with a learnable filter, as in Graph Convolutional Network (GCN) \cite{gcn} or based on the self-attention strategy which identifies the most important neighbors to be aggregated, as in Graph Attention Network (GAT) \cite{gat} or even based on a message-passing mechanism where features of both node in consideration and its neighboring nodes are combined to learn the local graph representation, as in \cite{message-passing}. While it presents an attractive opportunity for network neuroscience, only a few GNN architectures have so far been applied to the brain connectome and the most used GNN model is GCN \cite{gcn}. Therefore, we choose to cover it in detail in this section and for a more in-depth review of other variants of GNNs, we kindly refer the reader to \cite{Zhou2018,Wu2020,Zhang2020,Bronstein2017}. 

At the individual level, we define a graph as $G=( N, E, \mathbf{A}, \mathbf{F})$ representing a brain connectome, where $\mathbf{A}$ in $\mathbb{R}^{n\times n}$ is a connectivity matrix capturing the pairwise relationships between $n$ ROIs and $\mathbf{F}$ in $\mathbb{R}^{n\times f}$ is a feature matrix where $f$ is the dimension of node's feature. It is initialized using an identity matrix since in a brain graph there are no features naturally assigned to brain regions. At the population level, a graph can encode the relationship between a set of connectomes, where $\mathbf{A}$ denotes an affinity matrix capturing the similarity between $n$ brain graphs and $\mathbf{F}$ in $\mathbb{R}^{n\times f}$ is a feature matrix stacking the feature vector of size $f$ of each node (i.e, a single subject) in the  population graph (\textbf{Figure \ref{node-population}}). Therefore, $N$ is a set of nodes (i.e, ROIs or subjects), $E$ is a set of edges denoting either the biological connectivity between the brain regions or the similarity between brain graphs. In case the nodes do not have features, $\mathbf{F}$ can be represented by an identity matrix. Thus, we define the propagation rule of the GCN model as follows:

\begin{gather*}
 	 \mathbf{Z} = f_{\phi}(\mathbf{F}, \mathbf{A} \vert \mathbf{\Theta}) = \phi(\mathbf{\widetilde{D}}^{-\frac{1}{2}}\mathbf{\widetilde{\mathbf{A}}}\mathbf{\widetilde{D}}^{-\frac{1}{2}}\mathbf{F}\mathbf{\Theta})
 \label{eq:1} 
 \end{gather*}

$\mathbf{Z}$ is the learned graph representation resulting from a specific GCN layer. $\phi$ denotes the activation function such as Rectified Linear Unit (ReLU). Note that different activation functions can be used for different layers. $\mathbf{\Theta}$ is a filter denoting the graph convolutional weights. We define the graph convolution function as $f_{(.)}$ where $\mathbf{\widetilde{\mathbf{A}}} = \mathbf{\mathbf{A}} + \mathbf{I}$ with $\mathbf{I}$ being an identity matrix and $\mathbf{\widetilde{D}}_{ii} = \sum_{j}\mathbf{\widetilde{\mathbf{A}}_{ij}}$ is a diagonal matrix.

\begin{figure}
\centerline{\includegraphics[width=\columnwidth]{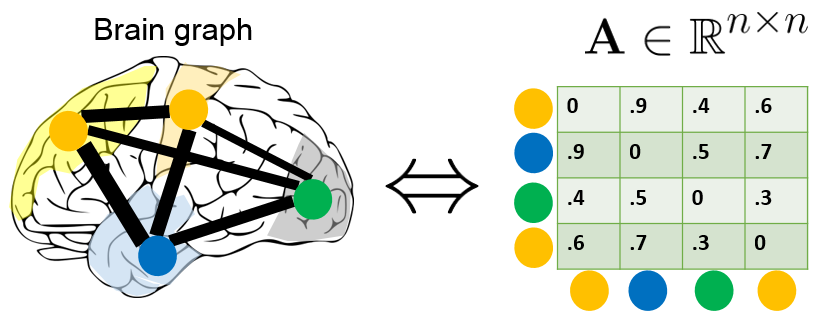}}\caption{Mathematical representation of a brain graph. Conventionally, a brain graph is represented by an adjacency matrix $\mathbf{A}\in\mathbb{R}^{n\times n}$ where $n$ is the number of brain regions extracted by the parcellation step and in this example $n=4$. Each element in this connectivity matrix encodes the weighted edges describing the connection existing between brain ROIs (nodes) \cite{Fornito2016}. }
\label{matrix}
\end{figure} 

\subsection{Brain graph overview}

To create a brain graph, one could consider the neurons and their synapses as the basic building blocks of graph (i.e., nodes and edges). However, this was demonstrated to be a computationally expensive task as it requires an intensive data acquisition and processing steps \cite{Fornito2015}. Hence, the scalable method of constructing a brain graph is to consider a set of neurons as a single node in the graph. This is achieved via several anatomical parcellation schemes applied to a particular imaging modality such as MRI (\textbf{Figure \ref{matrix}}). Conventionally, a connectome is an undirected graph which means that there is no inferences about possible directions of information flow between brain regions \cite{Fornito2015}. Therefore, a distinction is often drawn between three types of brain graphs: morphological brain graph, functional brain graph and structural brain graph \cite{Mahjoub2018,Logothetis2008,Booth2015}:
{\begin{itemize}

\item \emph{Morphological brain graphs}. Such graph were recently proposed in \cite{Mahjoub2018,Raeper2018,Soussia2018b,Lisowska2019}. This type of brain graphs involves cortical measures such as sulcal depth and cortical thickness to estimate the distance in morphology between brain regions. It is extracted from T1-weighted images which is first preprocessed using Freesurfer \cite{Freesurfer}. The preprocessing step mainly includes skull stripping, motion correction, T1-w intensity normalization, topology correction, segmentation of the sub-cortical white matter and deep grey matter volumetric structures and cortical hemisphere construction \cite{Dale1999}. Next, each hemisphere is parcellated into a set of brain regions using a particular atlas (e.g., Desikan-Killiany Atlas). First, the average value of a particular cortical attribute is computed for each ROI. The absolute difference between the average cortical attributes of a pair of brain regions is then computed which denotes the weighted edge linking two ROIs.

\item \emph{Functional brain graphs}. Conventionally, a functional graph is constructed from functional MRI (fMRI), more specifically from the blood-oxygen-level-dependent (BOLD) signal which shows the changes in blood oxygenation over time linked to neural activity \cite{Logothetis2008} in a particular region in the brain. First, the reported signal is averaged within each brain ROI. Next, a measure of correlation such as Pearson's correlation coefficient is computed between pairwise regions which results in the functional connectivity depicting the communication between pairs of brain regions. In functional brain graphs, nodes do not have features and edges are generally undirected and weighted.

\item \emph{Structural brain graphs}. A structural graph consists of a graph derived from diffusion tensor imaging (DTI) or diffusion spectrum imaging (DSI)  \cite{Booth2015,Hagmann2008,Basset2008}. Such neuroimaging data measure the diffusion of water molecules to generate contrast in MRI which allows for distinguishing gray matter from underlying white matter. There are several variants in generating structural connectomes among which we find computing the number of fibers connecting different brain regions and taking their absolute difference as a weight for structural connectivity. \cite{lerch2017studying}. 
\end{itemize}
}

\subsection{Literature search and taxonomy definition}

This review paper is both a position paper and a primer, promoting awareness and application of GNNs in network neuroscience along with giving a systematic taxonomy of existing models. Particularly, we focus on papers published between 1st of January 2017 to the 31th of December 2020. We have searched several electronic databases including IEEExplore, PubMed, Research Gate, Arxiv Sanity and Google Scholar using the following keywords: ``brain graph", ``brain network", ``connectome", ``GNN", ``graph representation learning", ``network neuroscience", ``graph neural networks". We collected 30 papers from the Medical Image Computing and Computer Assisted Intervention conference (MICCAI), the Information Processing in Medical Imaging (IPMI) conference, Journal of Neuroscience Methods, IEEE Transactions on Medical Imaging journal, Frontiers in Neuroscience, Neuroimage, Cerebral Cortex journals and bioRxiv. Non-GNN based papers related to brain graphs were excluded from our review such as those proposing CNN-based architectures for a connectomic related task or ML-based works such as \cite{Raeper2018,Lisowska2018,Lisowska2019}. We refer the reader to our GitHub link where all papers cited in our work are available at \url{https://github.com/basiralab/GNNs-in-Network-Neuroscience}. 

\begin{figure}[h]
\includegraphics[width=\columnwidth]{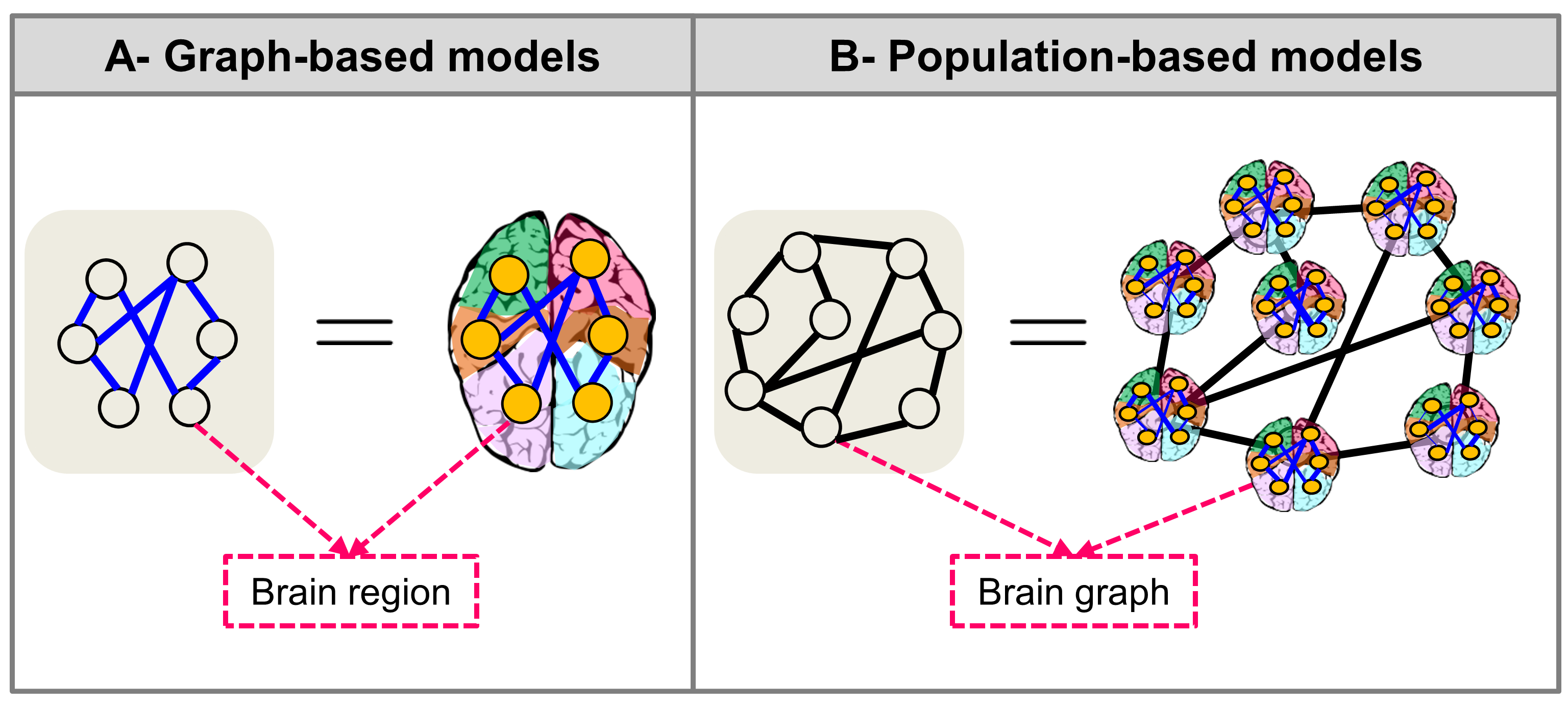}
\caption{GNN-based methods at a glance. Tow different families of existing models are depicted. (A) Graph-based models learn from a brain graph where nodes represent anatomical brain regions and edges denote the morphological, functional or structural connectivities. (B) Population-based models learn from a graph of subjects where nodes denote a brain graph of a particular subject and edges represent the similarity between nodes.}
\label{node-population}
\end{figure}

As shown in \textbf{Figure \ref{diagram}}, GNN-based network neuroscience comes in three different flavors: \emph{brain graph prediction, integration and classification}. Each of these flavors is further divided into three and two subcategories respectively, excluding the third one. Interestingly, we depict in our study two different groups of models: graph-based and population-based models. In the first group, designed frameworks learn from a brain graph where nodes represent anatomical brain regions and edges denote the morphological, functional or structural connectivities (\textbf{Figure \ref{node-population}}). Conversely, the second group harbours frameworks that take as input a graph of subjects where each node represents a single brain graph (i.e., subject) and each edge quantifies the pairwise similarity between two brain graphs (subjects).


\begin{figure*}
\centerline{\includegraphics[width=2\columnwidth]{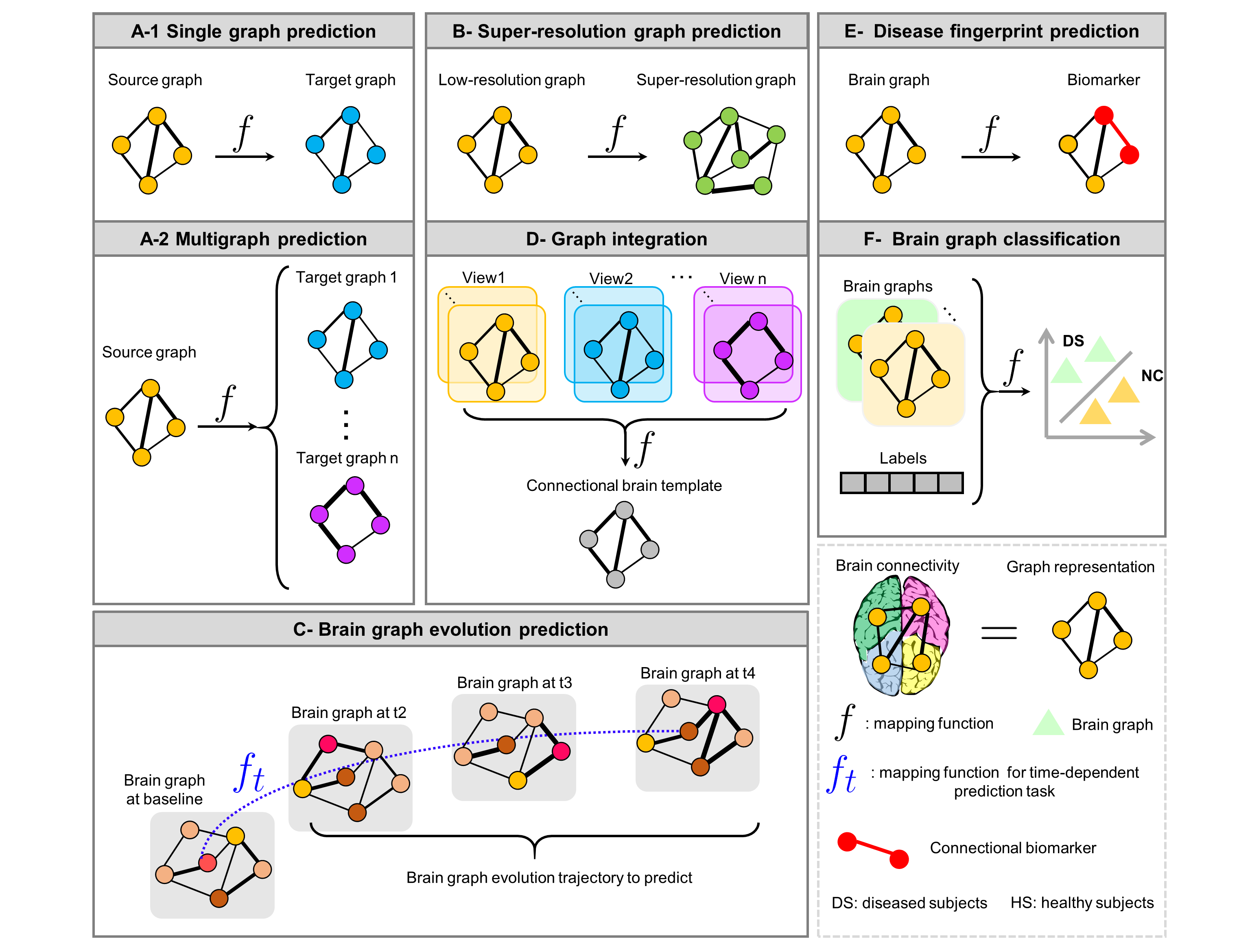}}
\caption{Schematic diagram of variants of network neuroscience-based goals achieved by existing GNN-methods. (A-1) Single graph prediction refers to the task of learning a mapping function $f$ for a cross-domain graph prediction purpose. (A-2) Multi-graph prediction, another subcategory of cross-domain graph prediction, refers to methods predicting multiple target graphs simultaneously from a single source graph. (B) Super-resolution prediction depicts the task of generating high-resolution graph with more nodes and connectivities given a low-resolution brain graph. (C) Graph integration refers to methods aiming to normalize a population of brain graphs by generating a representative connectional brain template. (F) Brain graph evolution prediction refers to the task of learning a mapping function $f_t$ that generates time-series brain graphs given a single brain graph acquired at baseline. (D) Disease fingerprint prediction is the task of discovering discriminative biomarkers existing in brain graphs. (E) Brain graph classification is the task of predicting the brain state of a subject (e.g., mild cognitive impairment (MCI) and Alzheimer's disease (AD)) either by learning on the whole brain graph or its learned embedding or even on a graph population. In the legend, we show an example of a brain graph which is an abstract graph representation of brain connectivities and we explain the abbreviations and mathematical symbols used in each block of the figure.
}
\label{taxonomy}
\end{figure*}

\subsection{Brain graph prediction}
Due to the high costs of medical scans, patients might have an MRI acquired at one timepoint and lack the follow-up MRI acquisition, or they might lack other scans such as diffusion MRI or resting-state functional MRI. Since brain graphs are derived from medical images, real-world connectomic datasets are usually incomplete. Furthermore, leveraging a small number of graphs for training learning-based models is unfeasible since it results in poor results \cite{LeCun2015}. Therefore, brain graph synthesis is important for boosting models designed for early disease diagnosis. We review in the three following subcategories all works belonging to the brain graph prediction flavor. This involves methods seeking to predict brain graphs across one of the three dimensions: domain, resolution, and time (\textbf{Figure \ref{diagram}}). The full list of brain graph prediction works is summarized in \textbf {Table \ref{prediction}} and descriptions for different loss terms are displayed in \textbf {Table \ref{tab1}} and \textbf {Table \ref{tab2}}. 

\subsubsection{Cross-domain graph prediction}
\hfill \break
\textbf{Problem statement.} Given a source graph $G_{S}$, our objective is to learn a mapping function $f: (\mathbf{A}_{S}, \mathbf{F}_{S})\mapsto (\mathbf{A}_{T}, \mathbf{F}_{T})$, which maps $G_{S}$ onto the target graph  $G_{T}$. We note that $G_{T}$ conventionally refers to a single target graph but it might refer to a tensor of target graphs which is the case of multi-graph prediction (\textbf{Figure \ref{taxonomy}-A-2}). 

\textbf{Existing works.} Existing GNN-based graph prediction frameworks lie into two groups: single graph prediction \cite{abu,Zhang2020,Zhang2020b,lgdada,hada,symdada} and multi-graph prediction tasks \cite{multigraphgan}. The main goal of the first group of works is to predict a target brain graph from a single source graph (\textbf{Figure \ref{taxonomy}-A-1}). For example, \cite{abu,Zhang2020,lgdada,hada,symdada} proposed geometric deep learning frameworks based on generative adversarial network (GAN) \cite{gan} to predict a target morphological brain graph from a source one and a structural brain graph from a functional one, respectively. Specifically, both frameworks built the generator and discriminator networks based on graph convolution network (GCN) \cite{gcn}. In addition to the GAN loss function, these works added new terms to strengthen the learning of the model. For instance, \cite{abu} proposed a topology-based loss term which enforces the alignment of the graph node strength distribution of the predicted target graphs and the ground truth ones, while \cite{Zhang2020} adopted a structure-preserving loss term to maximize the similarity between ground truth and predicted graphs from the generator. Other efforts to predict a target brain graph from a source one \cite{lgdada,hada,symdada} considered the domain alignment field as a solution for this problem which aims to match the statistical distribution of the source domain to the target one. For example, \cite{lgdada,symdada} leveraged two discriminators to align the training source and target brain graphs and adversarially regularize the training and testing source graphs embeddings. On the other hand, \cite{hada} proposed a hierarchical domain alignment strategy to match the distribution of the source graphs to the target graphs and simply used a single discriminator to regularize the whole framework. Other works adopted the autoencoder architecture to perform the graph prediction task. For instance, \cite{Zhang2020b} proposed a graph autoencoder based on a multi-stage graph convolution kernel. It represents the proposed propagation rule to the framework which involves the attention mechanism to perform the feature aggregation. Clearly, all these models perform one-to-one brain graph prediction. The shared aspect between these methods lies in solely performing one-to-one brain graph prediction. Using such models for predicting multiple target brain graphs from a source one, each model needs to be trained independently for each target domain. This hinders their scalability when predicting a multi-graph from a single source grain graph. Ideally, one would train a single model that simultaneously predicts a multi-graph from a single source brain graph such as synthesizing functional and structural brain graphs from a single morphological graph (\textbf{Figure \ref{taxonomy}-A-2}). To address this issue, a recent approach belonging to the multi-graph prediction group of works has been proposed and achieved better performance on morphological connectomic dataset \cite{multigraphgan}. It mainly consists in a GCN-based graph autoencoder including an encoder and a set of decoders all adversarially regularized by a single discriminator. Specifically, this work leveraged the data clustering and a topology conservation loss to avoid the mode collapse problem in GAN and preserve the brain graph structure, respectively. We provide a summary of all reviewed works in \textbf {Table \ref{prediction}}.



\textbf{Challenges and insights.}
There are a few challenges that need to be considered when predicting brain graphs. First, several frameworks are not built in an end-to-end fashion which may lead to accumulated errors across the learning steps. Second, for GAN-based frameworks, the mode collapse is an important issue to consider. However, a few are the works that handled this problem. Third, preserving brain graph topology is less considered in the prediction task. Devising topological-based convolutional functions may help improve graph representation learning. Fourth, we identified a single work in the category of multi-graph prediction which we consider more sophisticated for jointly synthesizing a set of brain graphs. Moreover, this model was evaluated only on morphological brain graphs. Further evaluations on different connectomic datasets might be valuable for clinical application. Finally, we identified graph-based \cite{abu,Zhang2020,Zhang2020b} and population-based models \cite{lgdada,hada,symdada,multigraphgan}. It would be better to have a denser explanation on the cases of using these two strategies and design a better topological properties conservation technique in the case of population-based models.

\subsubsection{Cross-resolution graph prediction}
\hfill \break
\textbf{Problem statement.} To generate a brain graph, one needs to pre-process and register the MRI data to a specific atlas space for automatic labelling of brain ROIs. This will result in a parcellation of the brain into $n$ anatomical brain regions defining the resolution of the graph \cite{parcellation1,parcellation2}. In other word, low-resolution and super-resolution brain graphs are generally derived from different MRI atlases which are the products of error-prone and time-consuming image processing pipelines \cite{parcellation2}. To circumvent the need for heavy image processing pipelines, prior works belonging to this subcategory of graph prediction aim to learn how to generate a high-resolution brain graph given a low-resolution one (\textbf{Figure \ref{taxonomy}-B}). Specifically, given a low-resolution brain graph $G_{l}=(N_{l}, E_{l}, \mathbf{A}_{l}, \mathbf{F}_{l})$, the goal is to learn a mapping function $f: (\mathbf{A}_{l}, \mathbf{F}_{l})\mapsto (\mathbf{A}_{s}, \mathbf{F}_{s})$, which maps $G_{l}$ onto the high-resolution graph $G_{s}=( N_{s}, E_{s}, \mathbf{A}_{s}, \mathbf{F}_{s})$, where $N_{l} < N_{s}$.

\textbf{Existing works.}
While several machine learning models have been proposed \cite{MhiriSuper,Cengiz2019}, the road to GNN-based super-resolution graph prediction is less traveled. \cite{Megi} conducted the first GNN-based work to generate a high-resolution brain graph from a single low-resolution one, with a detailed evaluation on real-world connectomic dataset. Inspired by the U-Net architecture, the authors proposed a graph U-autoencoder composed of a set of GCN layers and introduced a superresolving propagation rule based on grap eigendecomposition to generate the new connectivity matrix (i.e., brain graph). Their work demonstrated that the GNN-based model is a powerful tool to circumvent the time-consuming problem of medical image processing to estimate high-resolution brain graphs. The loss function proposed by \cite{Megi} can be found in \textbf {Table \ref{prediction}}. Such a multi-resolution representation of the brain connectivity provides a more holistic mapping of the brain as modular interactive system \cite{Fornito2016}.

\textbf{Challenges and insights.}
Future research should provide better GNN-based frameworks for superresolving brain graphs along with considering the computational efficiency of the model and the accurate prediction of the graph structure. One may provide a new propagation rule that superresolves a graph while considering the topological properties of the original graphs as well as addressing the domain shift (i.e., distribution change) that exists between the low- and super-resolution graph distributions. We note, also, that \cite{Megi} is solely designed for uni-resolution graph synthesis (i.e., mapping a source resolution onto a single target one) where the output of the model is a single super-resolution graph. Thus, future works might consider the case of multi-resolution graph synthesis where the trained model superresolves input brain graphs at multiple scales.


\subsubsection{Cross-time graph prediction}
\hfill \break
\textbf{Problem statement.}
Given a time-series brain graph data $G_{t_{i}}=(N, E, \mathbf{A}_{t_i}, \mathbf{F}_{t_i}), \ i \in \{1, \dots, n_t \}$ acquired at multiple timepoints, the goal is to learn a mapping function $f_t: (\mathbf{A}_{t_i}, \mathbf{F}_{t_i})\mapsto (\mathcal{T})$, which maps the baseline graph $G_{t_1}$ observed at the initial timepoint $t_1$ onto its evolution trajectory $\mathcal{T}=\{G_{t_{i+1}}\}_{i=2}^{n_t}$ that represents the follow-up brain graphs foreseeing the brain wiring evolution trajectory at future timepoints. 

\textbf{Existing works.}
Models that foresee the evolution of brain connectivity from a single baseline graph can be especially useful for both scientific discovery and clinical decision-making \cite{Ezzine,Ghribi}. Therefore, prior works proposed different GNN-based architectures that can be divided into two families: dichotomized learning-based models \cite{serkan,zeyneb} and end-to-end learning-based models \cite{evographnet}. These works aim to produce a trajectory either represented with one brain graph such as predicting brain connectivities of an Alzheimer's disease patient at 9-month after first scans or multiple follow-up brain graphs acquired at different timepoints (\textbf{Figure \ref{taxonomy}-C}). The first group of works relies on a population-based brain template which is the topic of the following section. Intuitively, they hypothesize that learning how a brain graph deviates from the brain template allows the identification of its most similar graphs. To do that, one can compute the similarity between brain graphs and the brain template. Ultimately, averaging the most similar graphs at a specific timepoint represents the predicted follow-up brain graph.


To do so, \cite{serkan} developed a GCN-based graph autoencoder adversarially regularized by a discriminator network that maps the brain graphs and the population template into a low-dimensional space. Next, a  computation of the residual between the graphs and the template embeddings is performed and finally the most similar graphs to the input one based on the resulting residual embeddings were selected. On the other hand, \cite{zeyneb} learns how to normalize a brain graph with respect to a graph population template by developing a GAN-based graph autoencoder where they leverage the edge-conditioned convolution (ECC) function which incorporates the edge weights into the convolution operation and performs pooling using a downsampling method based on eigenvector decomposition of the matrix.

This operation was designed to learn from graphs that lack node features which is the case of brain graphs studied in their work \cite{zeyneb}. Since these models were not designed in an end-to-end fashion, \cite{evographnet} introduced a unified brain graph evolution prediction framework. Specifically, a cascaded GAN-based architecture was designed where a generator synthesizing a brain graph at timepoint $t_i$ benefits from the previously synthesized graph at timepoint $t_{i-1}$ thereby foreseeing the full evolution trajectory of a brain graph. Similar to \cite{zeyneb}, edge convolution operations were leveraged for both generator and discriminator architectures. Reviewed works in this subsection are summarized in \textbf {Table \ref{prediction}}.

\textbf{Challenges and insights.}
Although \cite{serkan,zeyneb} methods can reliably predict the future brain connectivities, their performance is still less promising since their dichotomized learning strategy limits the scalability to predict jointly the follow-up brain graphs. Even though \cite{evographnet} is an end-to-end learning framework, it is still based on the edge convolution operation which hinders its scalability in terms of training time when applied on large-scale graphs. Essentially, such convolutional operation has been shown to be time-consuming when learned on large graphs \cite{NNConv} which hinders the capability of existing models in synthesizing brain longitudinal brain graphs. Therefore, more evolution trajectory prediction models that can help foster robust, scalable, and accurate prediction are highly desired. Ultimately, all these models overlook the synthesis of multiple trajectories representing brain graphs derived from different modalities. More advanced methods for jointly predicting evolution trajectories of multimodal brain graphs are compelling to design.

\begin{table*}[t!]
	\begin{center}
	\begin{threeparttable}
\caption{Brain graph prediction publications. In the first column, we list the name of existing works along with their citations for easy reference. In the second and third columns, we list the GNN model used and we briefly summarize all losses used to regularize the existing models. In the three last columns, symbols \cmark and \xmark denote whether the corresponding model is learned in an end-to-end fashion (E2E), learned from a brain graph (GM), or a population graph (PM), respectively. }

\label{prediction}

\begin{tabular}{|c |c |c |c |c |c |c|}
\hline
\textbf{Method} & \textbf{GNN}  & \textbf{Loss} & \textbf{E2E} & \textbf{GM} & \textbf{PM} \\\hline
CGTS\cite{abu} &GCN$^{\rm a}$ & \begin{tabular}[c]{@{}l@{}} $\mathcal{L}=\mathcal{L}_{gen1}(G_1,D_1)+\mathcal{L}_{gen2}(G_2,D_2)+$\\$\alpha_{1}[\mathcal{L}_{topo}(G_1,D_1)+\mathcal{L}_{topo}(G_2,D_2)]+$\\$\alpha_{2}\mathcal{L}_{cyc}(G_1,G_2)$ \end{tabular}& \cmark & \multirow{2}{*}{\cmark} & \multirow{2}{*}{\xmark} \\

\hline
MGCN-GAN\cite{Zhang2020} & GCN & $\mathcal{L}= \mathcal{L}_{GAN}+\alpha \mathcal{L}_{MSE} + \beta \mathcal{L}_{PCC}$ & \cmark & \cmark &\xmark \\ 

\hline
DMBN\cite{Zhang2020b}  &MGCK$^{\rm b}$ &$\mathcal{L}=\alpha \mathcal{L}_{global}+\beta \mathcal{L}_{local}+\mathcal{L}_{sup}$ & &\cmark &\xmark\\
\hline

HADA\cite{hada}  &GCN &\begin{tabular}[c]{@{}l@{}}$\mathcal{L}=\begin{cases} \mathcal{L}_{hier}=\mathbb{E}[log(D(\mathbf{Z}^{h-1}))]+\mathbb{E}[log(1-D(G^{h-1})]\\ \mathcal{L}_{source}=\mathbb{E}[log(D(\mathbf{Z}_{src}))]+\mathbb{E}[log(1-D(G)] \end{cases} $  \end{tabular}& \xmark &\xmark &\cmark\\
\hline

SymDADA\cite{symdada}  &GCN & \begin{tabular}[c]{@{}l@{}}$\mathcal{L}=\begin{cases} \mathcal{L}_{sym}=\mathbb{E}[log(D_{1}(\mathbf{Z}_{DA}))]+\mathbb{E}[log(1-D_{1}(G)]\\ \mathcal{L}_{dual1}=\mathbb{E}[log(D_{1}(\mathbf{Z}_{src}))]+\mathbb{E}[log(1-D_{1}(G)]\\
\mathcal{L}_{dual2}=\mathbb{E}[log(D_{2}(\mathbf{F}))]+\mathbb{E}[log(1-D_{2}(\hat{\mathbf{F}})]\end{cases} $  \end{tabular}& \xmark &\xmark &\cmark\\
\hline

LG-DADA\cite{lgdada}  &GCN & \begin{tabular}[c]{@{}l@{}}$\mathcal{L}=\begin{cases} \mathcal{L}_{align}= \mathbb{E}[log(D_{1}(\mathbf{F}_{s}))]+\mathbb{E}[log(1-D_{1}(\mathbf{Z}_{t})]\\ \mathcal{L}_{dual_{a}}=\mathbb{E}[log(D_{1}(\mathbf{Z}_{s}))]+\mathbb{E}[log(1-D_{1}(\mathbf{F}_{s})]\\
\mathcal{L}_{pred_{b}}=\mathbb{E}[log(D_{2}(\mathbf{F}_{t}))]+\mathbb{E}[log(1-D_{2}(\hat{\mathbf{F}_{t}})]\end{cases} $  \end{tabular}& \xmark &\xmark &\cmark\\
\hline

MultiGraphGAN\cite{multigraphgan} &GCN & \begin{tabular}[c]{@{}l@{}}$\mathcal{L}=\begin{cases} \mathcal{L}_{D}=\sum_{j=1}^{c}(\mathcal{L}^{j}_{WD}+\alpha_{1} \mathcal{L}^{j}_{gdc}+\alpha_{2} \mathcal{L}^{j}_{gp}) \\\mathcal{L}_{G}=\sum_{j=1}^{c}(\mathcal{L}^{j}_{gen3}+\alpha_{1} \mathcal{L}^{j}_{topo2}+\alpha_{2} \mathcal{L}^{j}_{rec}+\alpha_{3} \mathcal{L}^{j}_{info})
\end{cases} $ \end{tabular} & \xmark &\xmark &\cmark\\
\hline

GSR-Net\cite{Megi} &GCN &\begin{tabular}[c]{@{}l@{}}$\mathcal{L}= \mathcal{L}_{hr}+ \mathcal{L}_{eig} + \alpha \mathcal{L}_{rec}$   \end{tabular}& \cmark & \cmark &\xmark \\
\hline

RESNet\cite{serkan}  &GCN &\begin{tabular}[c]{@{}l@{}}$\mathcal{L}_{emb}= \mathbb{E}[log(D_{1}(\mathbf{A}_{t_0}))]+\mathbb{E}[log(1-D_{1}(\mathbf{z}_{t_0})]$   \end{tabular} & \xmark &\cmark &\xmark\\
\hline

gGAN\cite{zeyneb}  &ECC$^{\rm c}$ &\begin{tabular}[c]{@{}l@{}}$\mathcal{L}=\mathcal{L}_{adv}+\alpha \mathcal{L}_1 (N)$ \end{tabular} & \xmark &\cmark &\xmark\\
\hline

EvoGraphNet\cite{evographnet}  &ECC &\begin{tabular}[c]{@{}l@{}} $\mathcal{L}= \sum_{i=1}^{m}(\alpha_{1}\mathcal{L}_{adv}(Gi,Di) + \frac{\alpha_{2}}{n_s}\sum_{s=1}^{n_s}\mathcal{L}_{1}(G_i,s) + \frac{\alpha_{3}}{n_s}\sum_{s=1}^{n_s}\mathcal{L}_{KL}(t_i,s)$\end{tabular} &\cmark &\cmark &\xmark\\
\hline
\end{tabular}

\begin{tablenotes}
\item $^{\rm a}$Graph convolutional networks; $^{\rm b}$Multi-stage graph convolution kernel; $^{\rm c}$Edge-conditioned convolution
\end{tablenotes}
\end{threeparttable}
\end{center}
\end{table*}

\subsection{Brain graph integration}
\textbf{Problem statement.}
Consider the case where a subject is represented by a set of brain graphs derived from the same modality $\mathcal{G^s}=\{G^{m_i}\}_{m_i=1}^{n_m}$ where $G^{m_i}=( N, E, \mathbf{A}, \mathbf{F})$, $n_m$ is the number of modalities and $i \in \{1, \dots, n_m \}$. The goal of this category of models is to learn a mapping function $f: \{\mathcal{G}^{i}\}_{i=1}^{n_s} \mapsto \mathbf{C}$, which integrates $\{\mathcal{G}^{i}\}_{i=1}^{n_s}$ the multimodal brain graphs of all subjects in a given population into a single holistic fingerprint brain graph $\mathbf{C}$ that represents the shared common connectivity patterns across subjects. Such a representative connectional brain template called CBT is important for spotting  biological patterns that alter when comparing typical to atypical populations.


\textbf{Existing works.} \cite{ugur} proposed a clustering-based GNN model that produces a brain template given a set of graphs (\textbf{Figure \ref{taxonomy}-D}). After clustering the population into groups, the proposed MGINet architecture basically integrates the different graphs that represent subjects belonging to the same cluster into a single graph by extracting the most representative edges linking two nodes that represent meta-paths of the graph. Next, a learning averaging method is leveraged to produce the final CBT of the population. While effective, this model is not trained in an end-to-end fashion since the clustering, cluster-specific CBT generation, and the population CBT estimation blocks are learned disjointly. Therefore, cumulated errors across these blocks might produce a less centered brain template. More recently, \cite{DGN} proposed Deep Graph Normalizer (DGN), a GNN-based model that integrates a population of multiview brain graphs named ``multigraph" into a connectional brain template (CBT) in an end-to-end learning fashion. Specifically, a multigraph encodes different facets of the brain where the exisiting connection between two nodes is encoded in a set of edges of multiple types. Each edge type denotes a particular form of brain connectivity derived from a particular neuroimage data. DGN was proposed to integrate the complementary information across all subjects of the population by (i) training a set of edge conditioned convolutional layers each learns the embeddings of brain regions in the graph, (ii) introduce a series of tensor operations to calculate the pairwise absolute difference of each pair of the learned node embeddings which results in a subject-based CBT, and finally, (iii) estimate the CBT of the population by selecting the median of all subject-based CBTs.  All loss functions proposed in the aforementioned works are summarized in \textbf {Table \ref{cbt}}.

\begin{table*}[h]
	\begin{center}
\begin{threeparttable}
\caption{Brain graph integration publications. A brief description of the GNN, loss functions training manner and data representations used in each paper. Symbols \cmark and \xmark denote whether the corresponding model is learned in an end-to-end fashion (E2E), learned from a brain graph (GM), or a population graph (PM), respectively.}
\label{cbt}

\begin{tabular}{|c |c |c |c |c |c |c|}
\hline
\textbf{Method} & \textbf{GNN}  & \textbf{Loss}  & \textbf{E2E} & \textbf{GM} & \textbf{PM} \\ \hline
MGINet\cite{ugur} &GCN$^{\rm a}$ & \begin{tabular}[c]{@{}l@{}}
$\mathcal{L}=\underbrace{\frac{1}{n_c}\sum_{s=1}^{n_c}\sum_{k=1}^{p}\parallel \mathbf{C}_{s}-\mathbf{A}^{s}_{k}\parallel^{2}}_{\mathcal{L}_{sub}}$ \\$+  \underbrace{\sum_{s=1}^{n_c}\sum_{k=1}^{p}\parallel \mathbf{Z}_{c}-\mathbf{A}^{s}_{k}\parallel^{2}}_{\mathcal{L}_{clt}}$
\end{tabular}&  \xmark &\cmark &\xmark \\ \hline

DGN\cite{DGN}  &ECC$^{\rm b}$ &\begin{tabular}[c]{@{}l@{}}
$\mathcal{L}= \frac{1}{\vert T \vert}\sum_{s=1}^{\vert T \vert} \underbrace{(\lambda\sum_{v=1}^{n_m} \sum_{s=1}^{n_s}\parallel \mathbf{C}_{s}-\mathbf{T}_{i}^{v}\parallel_{F})}_{\mathcal{L}_{SNL_{s}}}$\end{tabular} &\cmark &\cmark &\xmark\\
\hline
\end{tabular}

\begin{tablenotes}
\item $^{\rm a}$Graph convolutional networks; $^{\rm b}$Edge-conditioned convolution
\end{tablenotes}
\end{threeparttable}
\end{center}
\end{table*}

\textbf{Challenges and insights.}
While offering a tool to distinguish between healthy and disordered populations, the proposed GNN-based CBT learning architectures are limited to some extent. First, they solely generate CBTs of a population of multimodal \emph{fixed-size} brain graphs (i.e., number of nodes and edges are the same across all graphs) which hinders their integration capability when subjects are represented by non-isomorphic (i.e., multi-resolution) brain graphs with varying topologies. Second, they do not explain which brain graph connectivity types contributes most to the CBT learning process. Seminal works on GNN explainability including GNNExplainer and GraphLIME \cite{gnnexplainer, graphlime} aimed to explain a graph by extracting the crucial features in the graph that affect the prediction of the GNN. The rationale behind the explanation of brain graph integration models can be motivated from a neuroscience perspective: different brain connectivity types might be more representative of the brain wiring than others. Besides, there is evidence that the sub-graphs identified by the explainer model represent the shared connectivities in a population that are integrated by the GNN model. Indeed, the identification of representative modules in the graph might help reduce the errors made by integrator models. Therefore, considering the explanation of GNN-based multigraph integration models needs to be developed to optimize the connectional template estimation and propel the field of disease prediction. 

\subsection{Brain graph classification}
Perhaps the first and most fundamental question that one can ask about a brain graph is whether it is useful for early diagnosing a disease. A straightforward way is to train a classifier to distinguish between healthy and unhealthy subjects and to build a model that identifies the most discriminative connectivities in the brain that fingerprints a typical disease. For this reason, we divide the brain graph classification flavor into two categories: brain state classification and biomarker identification (\textbf{Figure \ref{taxonomy}-E-F}). The first one asks how to classify brain states such as predicting if a subject is disordered or healthy. This category gives rise to three other groups (\textbf{Figure \ref{diagram}} orange boxes top): graph embedding-based, graph-based and population-based classification methods. The first one depicts models classifying brain states using the learned brain graph embedding. In the second one, models learn to classify subjects using the whole brain graph where a node denotes a brain region. Rather than learning on a whole-brain graph, methods belonging to the third group performs brain state classification by learning on graph population where a node denotes a subject's brain graph. The second category asks how to identify discriminative biomarkers that have the potential to be used in clinical settings to characterize disease-related brain connectivities in patients and treatment response. We provide a summary of different disease classification works in \textbf {Table \ref{classification}}.

\subsubsection{Brain state classification}
\hfill \break
\textbf{A- Graph embedding-based classification}
\hfill \break

\textbf{Problem statement.}
Given a brain graph $G=(N, E, \mathbf{A}, \mathbf{F})$, the goal is to learn the compressed low-representation of the graph that is the embedding $\mathbf{Z}$ and leverage it to learn a mapping function $f: \mathbf{Z} \mapsto y$, which maps the input graph embedding $\mathbf{Z}$ onto its label $y$ (e.g., normal control or autistic).

\textbf{Existing works.}
\cite{ACE} designed a GAN-based approach where the generator is a set of GCN layers learning the deep low-dimensional representation of the input graph. Next, the graph embeddings of all graphs are passed to a linear support vector machine (SVM) to classify connectomes of late mild cognitive impairment and Alzheimer's disease patients. Leveraging the brain graph embedding to predict the clinical depression scores of patients were further designed in \cite{ipmi1}. Assuming that a brain graph does not capture the high-order relation between brain regions, \cite{HCAE,mert} introduced a hyperconnectome where a node can be linked to more than one node. Specifically, \cite{HCAE} learned the embedding of a multi-view brain hyperconnectome using a hypergraph neural network (HGNN) encoder. In that way, they classify the brain state using different brain modalities. Instead of considering an ROI as a node in the hypergraph, \cite{mert}  designed a hypergraph of subjects where a brain graph of a patient represents a node. The authors proposed a new hypergraph pooling-unpooling layer to learn embedding capturing the relationships between subjects then leveraged it to predict their corresponding brain states. Other works preferred to learn the graph embedding using graph attention network (GAT) which focuses on learning from the most relevant nodes in the graph. For instance, \cite{tmi} proposed to combine the structural and functional brain graphs into a single graph representation. Specifically, they learn the embedding of a single graph that is represented by an adjacency matrix denoting the structural connectivities and a feature matrix denoting the functional connectivities. The resulting learned latent representation was passed to a multi-layer perceptron classifier to predict frontal lobe epilepsy, temporal lobe epilepsy, and healthy subjects. All the aforementioned models lack interpretation of the captured features in the embedding space which can help better understand the original connectivity pattern that yields the illness. Therefore, \cite{xiao} another attention-based network, proposed a combination of different GNN layers (Edge-Weighted GAT layer, followed by Diffpool layers) to learn the graph embedding and identify the Bipolar Disorder patients. Essentially, they performed a visual interpretation of the attention map generated by the edge-weighted GAT network so that they detect the particular associations of abnormalities in the functional brain graph.

\textbf{Challenges and insights.}
Building models that not only predict the brain state of the subject but also explaining why it makes such prediction, as well as integrating all types of brain graphs in the learning process, is a compelling research direction that remains unexplored.

\hfill \break
\textbf{B-Graph-based classification}
\hfill \break

\textbf{Problem statement.}
While the previous group of works performed the classification of subject using the learned graph embeddings, works belonging to the graph-based classification group leverage the whole brain graph to make the target prediction.

\textbf{Existing works.}
Recently, several pioneering models have been devised to predict brain disease by learning on brain graphs. For instance, \cite{class1} proposed a combination of GCN model along with long short term memory (LSTM) network to classify functional connectivity of demented and healthy subjects. They included the prediction of gender and age in a regularization task to improve the disease classification. \cite{class2,class3} proposed another combination of   GCN and recurrent neural network (RNN) models which mainly deal with both brain structural and functional connectivities to identify the mild cognitive impairment patients. Later on, \cite{class4} leveraged siamese graph convolutional neural network (s-GCN) to learn the similarity between a pair of graphs and incorporated the learned similarity into the classification step. From another perspective, \cite{class5} assume that detecting the brain disease state is relative to study the brain region's function that can be represented by its multi-hops connectivity profile. Therefore, a GNN model taking as input the functional brain graph learns how many levels of nearest neighbors of a specific ROI that need to be considered in a brain graph classification task.

\textbf{Challenges and insights.}
Despite the effectiveness of these models, a common limitation is their difficulty to explain the classification results in a neuroscientifically explainable way. Future developments might include an explainable study of how GNN-based models are learned in a brain graph classification task.

\hfill \break
\textbf{C-Population-based classification}
\hfill \break

\textbf{Problem statement.}
Owing to the fact that different types of data (i.e., images, phonetype, connectomes, genetic sequence) provide complementary information for analyzing neurological disorders, works under this group of brain state classification aim to construct a population graph describing the relationship between subjects. Let $P=(N_p, E_p, \mathbf{A}_p, \mathbf{F}_p)$ be the graph population where $N_p$ is a set of nodes (i.e, brain graphs), $E_p$ is a set of edges denoting either the correlation or similarity between the nodes using the brain connectivities or non-connectomic data, $\mathbf{A}_p$ in $\mathbb{R}^{n_s\times n_s}$ is an adjacency matrix where $n_s$ is the number of subjects, and $\mathbf{F}_p$ in $\mathbb{R}^{n_s\times f}$ is a feature matrix stacking the feature vector (i.e., connectomic or non-connectomic features) of nodes in the graph. The goal is to learn a mapping function $f: P \mapsto y$, which maps $P$ onto its label (e.g., normal control or autistic).

\textbf{Existing works.}
Several works have been proposed in the state-of-the-art to early identify the brain state of a subject by creating a population graph \cite{pop1,ipmi2}. They mainly constructed the graph using imaging and non-imaging data where nodes are subjects represented by neuroimaging data and edges denote the pairwise similarity between phenotypic data of the subjects, then a GCN-based classifier learned from it to predict the node label (i.e, a brain graph). While such population-based classification works provide powerful frameworks for healthy and disordered population analysis and a combination of imaging of non-imaging data, we identified a single work where nodes in the graph denote connectomes of the subjects \cite{pop2}. Specifically, this model aims to integrate both functional and structural brain graphs to early predict Alzheimer's disease. They introduced a calibration technique to generate the adjacency matrix representing the population. Next, they built two different GCNs each trained on a specific modality, and generate a clinical score of the subjects in the population.

\textbf{Challenges and insights.}
Although the unique population-based work that we found for brain graph classification achieved promising results on a relatively large dataset, existing connectomic datasets are still scarce and have missing instances. Such a case represents one of the main difficulties encountered in building classification models as they tend to overfit. Devising GNN-based models that from minimal resources they synthesize brain graphs along with performing a classification task is highly needed.

\begin{table*}[h]
	\begin{center}
\begin{threeparttable}
\caption{Disease classification publications. In the first column, we list the name of existing works along with their citations for easy reference. In the following three columns, we list the GNN models, losses for reguarization and the classifier used in different works. In the three last columns, symbols \cmark and \xmark denote whether the corresponding model is learned in an end-to-end fashion (E2E), learned from a brain graph (GM) or from a population graph (PM), respectively. }
\label{classification}
\begin{tabular}{|c |c |c |c |c |c |c |c|}
\hline
\textbf{Method} & \textbf{GNN}  & \textbf{Loss} & \textbf{Classifier}& \textbf{E2E} & \textbf{GM} & \textbf{PM} \\\hline
ACE\cite{ACE} &GCN & \begin{tabular}[c]{@{}l@{}} $\mathcal{L}=\mathbb{E}[log(D(\mathbf{Z}))]+\mathbb{E}[log(1-D(G(\mathbf{F},\mathbf{A})]$ \end{tabular}& SVM$^{\rm a}$& \xmark & \cmark & \xmark \\\hline

HCAE\cite{HCAE}& HGNN$^{\rm b}$ & \begin{tabular}[c]{@{}l@{}} $\mathcal{L}=\mathbb{E}[log(D(\mathbf{Z}))]+\mathbb{E}[log(1-D(G(\mathbb{F},\mathbb{A})]$ \end{tabular}& SVM& \xmark & \cmark & \xmark \\\hline


ADB-NN\cite{tmi}&SA$^{\rm c}$ & \begin{tabular}[c]{@{}l@{}} $\mathcal{L}= - \sum{m} y_{m} log(softmax(f(\mathbf{W}^{d} \odot \mathbf{B})$ \end{tabular}& FC$^{\rm d}$ & \cmark & \xmark & \cmark \\\hline 

Blind-GCN\cite{ipmi1}&GCN & \begin{tabular}[c]{@{}l@{}} $\mathcal{L}= \parallel \hat{Y} - Y \parallel_{F} + \alpha_{1}\parallel \mathbf{W} \parallel_{2} + \alpha_{2}\parallel \hat{H} - H \parallel $ \end{tabular}& MLP$^{\rm e}$ & \cmark & \xmark & \cmark \\\hline  

s-GCN\cite{class1}&GCN & \begin{tabular}[c]{@{}l@{}} $\mathcal{L}=\max (0, \sigma^{2+} - a) + \max(0,m-(\mu^{+}-\mu^{-}))$ \end{tabular}& k-NN$^{\rm f}$ & \cmark & \xmark & \cmark \\\hline 

GC-LSTM\& DS-GCN\cite{class2,class4}&GCN & \begin{tabular}[c]{@{}l@{}} $\mathcal{L}=\alpha_{1}\mathcal{L}_{dis} + \alpha_{2}\mathcal{L}_{gen} + \alpha_{3}\mathcal{L}_{a}$ \end{tabular}& FC & \cmark & \xmark & \cmark \\\hline 

Population-method1$^{\rm g}$\cite{pop1}&GCN & \begin{tabular}[c]{@{}l@{}} $\mathcal{L}= -\sum_{s} p(y^{s})\log q(y^{s})$ \end{tabular}& FC & \cmark & \xmark & \cmark \\\hline 

\begin{tabular}[c]{@{}l@{}}Population-method2$^{\rm g}$,\\ InceptionGCN\cite{pop1,ipmi2}\end{tabular}&GCN & \begin{tabular}[c]{@{}l@{}} $\mathcal{L}= -\sum_{s} p(G^{s})\log q(G^{s})$ \end{tabular}& FC & \cmark & \xmark & \cmark \\\hline 

\hline
\end{tabular}
\begin{tablenotes}
\item $^{\rm a}$Support vector machine; $^{\rm b}$Hypergraph convolution; $^{\rm c}$Self attention mechanism; $^{\rm d}$Fully connected layer; $^{\rm e}$k-nearest neighbour; MLP$^{\rm f}$Multi layer perceptron; $^{\rm g}$ we named these methods since they were not named in the original papers.
\end{tablenotes}
\end{threeparttable}
\end{center}
\end{table*}

\subsubsection{Biomarker identification}

\hfill \break
\textbf{Problem statement.}
A complementary but distinct line of work capitalizes on the emerging field of interpretability of GNN intending to develop models that identify the most discriminative biomarkers in addition to predicting the label of brain graphs.

\textbf{Existing works.}
\cite{bio1} proposed a GNN classifier that sequentially stacks a set of Message Passing Neural Networks (NNconv) and pooling layers to estimate the class of the functional brain graphs and generate a set of sub-graphs (i.e., one sub-graph is a set of ROIs) which represent the identified biomarkers. Next, they interpret the importance of each sub-graph identified by performing an extensive comparison between their GNN and existing classifiers. On the other hand, \cite{bio2,bio4} proposed a GAT-based architecture to predict the most discriminative ROIs for identifying a disease. They specifically proposed a pooling layer along with a regularization loss term to soften the distribution of the node pooling scores generated by the network. Detailed loss functions are included in \textbf {Table \ref{biomarker}}.

\textbf{Challenges and insights.}
For the above mentioned works, we consider two challenges: reproducibility and explainability.\begin{marginnote}
{Graph attention convolution propagation rule}{Introduced in \cite{gat}, it includes the attention mechanism to propagate signals between nodes $i$ and $j$: $\mathbf{h}^{\prime}_{i} = \sigma ( \sum_{j\in \mathcal{N}_{i}} \alpha_{ij} \mathbf{W}\mathbf{h}_{j})$. $\sigma$ is an activation function, $\alpha_{ij}$ is a weighting factor denoting the importance of a node $j$ to the node $i$, $\mathcal{N}_{i}$ is a set of neighborhood of node $i$ in the graph and $\mathbf{W}\in \mathbb{R}^{f\times f}$ is a linear transformation weight matrix where $f$ is the feature dimension of a node.}
\end{marginnote} 

The first limitation lies in ensuring that the proposed GNN-based model is reproducible in identifying the discriminative biomarkers. Ideally, one would design a model that reproduces similar results when leveraging different training strategies such as cross-validation or few shot learning. This is of great interest since it will help understand which model to trust in the network neuroscience field. The second limitation implies that these GNN-based models are explanation agnostic for the target prediction task. Yet, understanding the learning process of the model can increase the trustworthiness of the model and identify the strength and falls of the model when applied in real-world applications. Therefore, for a better understanding of neurological disorders, it would be beneficial to design explainable models that take as input the learning model and its prediction and try to identify the most important feature that is crucial for the task at hand.


\begin{table*}[h]
	\begin{center}
\begin{threeparttable}
\caption{Biomarker identification publications. A brief description of the GNN, loss functions training manner and data representations used in each paper.}
\label{biomarker}
\begin{tabular}{|c |c |c |c |c |c |c|}
\hline
\textbf{Method} & \textbf{GNN}  & \textbf{Loss}  & \textbf{E2E} & \textbf{GM} & \textbf{PM} \\\hline
\cite{bio1} &NNconv$^{\rm a}$ & \begin{tabular}[c]{@{}l@{}}$\begin{cases}\mathcal{L}_{reg1}=\alpha \sum_{l=1}^{L} ( \parallel \mathbf{w}^{(l)} \parallel_{2} - 1)^{2}\\ I_{1} = \frac{1}{S} \sum_{s} tanh(log_{2}(p(c=c_{s}|\mathbf{G}_{sj})/(1-p(c=c_{s}|\mathbf{G}_{sj})))) \end{cases}$
\end{tabular}&  \xmark &\cmark &\xmark \\ \hline

PR-GNN\cite{bio2}  & GATConv$^{\rm b}$ &\begin{tabular}[c]{@{}l@{}}
$\mathcal{L}= \mathcal{L}_{ce} + \alpha_{1}\sum_{l=1}^{L} \mathcal{L}_{dist}^{(l)} + \alpha_{2}\sum_{c=1}^{C} \mathcal{L}_{gdc}^{c}$\end{tabular} &\cmark &\cmark &\xmark\\
\hline

EGAT\cite{bio3}  & GATConv$^{\rm b}$ &\begin{tabular}[c]{@{}l@{}} $\begin{cases} \mathcal{L}_{reg2}= \parallel \mathbf{E},\mathbf{SS}^{T} \parallel_{F}\\ I_{2} = \frac{ \alpha ((\mathbf{a}^{p})^{T}  (W^{p}\mathbf{x}_{i} \parallel W^{p} \mathbf{x}_{j}))}{\alpha (\mathbf{x}_{i}, \mathbf{x}_{j})} \end{cases}$
\end{tabular} &\cmark &\cmark &\xmark\\\hline

BrainGNN\cite{bio4}  & Ra-GNN$^{\rm c}$ &\begin{tabular}[c]{@{}l@{}} $\mathcal{L}= \mathcal{L}_{ce} + \alpha_{1}^{l}\sum_{l=1}^{L} \mathcal{L}_{reg}^{(l)} + \alpha_{2}^{l}\sum_{l=1}^{L} \mathcal{L}_{TPK}^{(l)} +  \alpha_{3}^{l} \mathcal{L}_{gdc}$
\end{tabular} &\cmark &\cmark &\xmark\\\hline

\end{tabular}
\begin{tablenotes}
\item $^{\rm a}$Message Passing Neural Networks; $^{\rm b}$ Graph Attention Convolution; $^{\rm c}$ ROI-aware graph convolutional neural network
\end{tablenotes}
\end{threeparttable}
\end{center}
\end{table*}

\begin{table}[b!]

\caption{A brief summary of different losses used in each paper reviewed in this work (Part 1). The first column stands for loss terms included in \textbf {Table \ref{prediction}}, \textbf {Table \ref{cbt}}, \textbf {Table \ref{classification}} and \textbf {Table \ref{biomarker}}. The second column describes the meaning of each term in order to understand the full loss function included in the previous tables.}
\label{tab1}
\begin{center}
\setlength{\tabcolsep}{3pt}
\begin{tabular}{|p{50pt}|p{190pt}|}
\hline
\textbf{Losses}  & \textbf{Remark} \\\hline
$\mathcal{L}_{gen}(G_i,D_i)$ & \begin{tabular}[c]{p{190pt}}Adversarial loss introduced in \cite{gan} to optimize the generator $G_i$ and discriminator $D_i$ and the absolute difference between the output of the generators and the ground truth where $i \in \{1,2\}$.\end{tabular}\\\hline

$\mathcal{L}_{topo1}(G_i,D_i)$ & \begin{tabular}[c]{p{190pt}} Topology-based loss aims to preserve the node strength of the ground truth brain graphs
\end{tabular}\\\hline

$\mathcal{L}_{cyc}(G_1,G_2)$ & \begin{tabular}[c]{p{190pt}} Topology-based cycle consistency loss that ensures the accurate mapping of the target to the source graph. 
\end{tabular}\\\hline

$\mathcal{L}_{hier}$ & \begin{tabular}[c]{p{190pt}} Domain alignment loss function for hierarchical domain alignment of source to target brain graphs and $\mathcal{L}_{source}$ is the GAN loss function for learning source graph embeddings.
\end{tabular}\\\hline

$\mathcal{L}_{GAN}$ & \begin{tabular}[c]{p{190pt}} is the adversarial loss, $\mathcal{L}_{MSE}$ and $\mathcal{L}_{PCC}$ are the element-wise loss and similarity maximization loss to ensure the similarity between the ground truth and predicted graph.
\end{tabular}\\\hline

$\mathcal{L}_{global}$ & \begin{tabular}[c]{p{190pt}} Edge-based topological loss that ensures the edge reconstruction in the target brain graph, $\mathcal{L}_{local}$ aims to preserve the local connectivity in the graph, $\mathcal{L}_{sup}$ is the supervised loss of generating brain saliency maps.
\end{tabular}\\\hline

$\mathcal{L}_{sym}$ & \begin{tabular}[c]{p{190pt}} Domain alignment loss that enforces the symmetric domain adaptation of the source brain graphs to the target ones and the target graphs to the source ones.
\end{tabular}\\\hline

$\mathcal{L}_{dual1}$ & \begin{tabular}[c]{p{190pt}} First part of dual adversarial regularization of source brain graph embedding which aligns the learned embeddings to the real source distribution.
\end{tabular}\\\hline

$\mathcal{L}_{dual2}$ & \begin{tabular}[c]{p{190pt}} Second part of dual adversarial regularization of the source domain which aligns the source distribution to the target one by optimize the target brain graph prediction. 
\end{tabular}\\\hline

$\mathcal{L}_{D}$ & \begin{tabular}[c]{p{190pt}} Discriminator loss composed by three loss terms: $\mathcal{L}^{j}_{WD}$, $\mathcal{L}^{j}_{gdc}$ and $\mathcal{L}^{j}_{gp}$.
\end{tabular}\\\hline

$\mathcal{L}^{j}_{WD}$ & \begin{tabular}[c]{p{190pt}} Adversarial loss of the discriminator based on the Wasserstein distance.
\end{tabular}\\\hline

$\mathcal{L}^{j}_{gdc}$ & \begin{tabular}[c]{p{190pt}} Graph domain classification loss computed for a specific cluster $j$ which is the mean squared loss between the ground truth and generated target graph labels. 
\end{tabular}\\\hline

$\mathcal{L}^{j}_{gp}$ & \begin{tabular}[c]{p{190pt}} Gradient penalty loss term used to improve  the  training  stability  of  the  model
\end{tabular}\\\hline

$\mathcal{L}_{G}$ & \begin{tabular}[c]{p{190pt}} Generator loss composed by four loss terms: $\mathcal{L}^{j}_{gen3}$, $\mathcal{L}^{j}_{topo2}$, $\mathcal{L}^{j}_{rec}$ and $\mathcal{L}^{j}_{info}$.
\end{tabular}\\\hline

$\mathcal{L}^{j}_{gen3}$ & \begin{tabular}[c]{p{190pt}} Adversarial loss term used to train the generator network of the cluster $j$.
\end{tabular}\\\hline

$\mathcal{L}^{j}_{topo2}$ & \begin{tabular}[c]{p{190pt}}Topology-based loss which enforces the generator $j$ to preserve the centrality scores of nodes in the target graph as well as learning the global graph structure.
\end{tabular}\\\hline

$\mathcal{L}^{j}_{rec}$ & \begin{tabular}[c]{p{190pt}} Topology-based reconstruction loss that guarantees that the predicted target brain graphs can be used to recreate the source graph. 
\end{tabular}\\\hline

$\mathcal{L}^{j}_{info}$ & \begin{tabular}[c]{p{190pt}}
Information maximization loss term to ensure that the predicted graphs are correlated when predicted from the generator $j$.  
\end{tabular}\\\hline

\hline
\end{tabular}
\end{center}
\end{table}

\begin{table}[t!]
\caption{A brief summary of different losses used in each paper reviewed in this work (Part 1). The first column stands for loss terms included in  \textbf {Table \ref{prediction}}, \textbf {Table \ref{cbt}}, \textbf {Table \ref{classification}} and \textbf {Table \ref{biomarker}}. The second column describes the meaning of each term in order to understand the full loss function included in the previous tables.}
\label{tab2}
\begin{center}
\setlength{\tabcolsep}{3pt}
\begin{tabular}{|p{25pt}|p{210pt}|}
\hline
\textbf{Losses}  & \textbf{Remark} \\\hline
$\mathcal{L}_{hr}$& \begin{tabular}[c]{p{200pt}}
Super-resolution loss term aims to minimize the MSE between the ground truth and predicted super-resolution brain graphs.  
\end{tabular}\\\hline

$\mathcal{L}_{eig}$& \begin{tabular}[c]{p{200pt}}
Eigen-decomposition loss enforcing the eigen-decomposition of the predicted super-resolution brain graph to match the one of the ground truth graph.
\end{tabular}\\\hline

$\mathcal{L}_{rec}$& \begin{tabular}[c]{p{200pt}}
Reconstruction regularization loss term which is the MSE between the ground truth and the predicted node feature embedding of the low-resolution brain graphs. \end{tabular}\\\hline

$\mathcal{L}_{emb}$& \begin{tabular}[c]{p{200pt}}
Adversarial loss for regularizing the brain graph embeddings at the first timepoint $t_0$. It matches the distribution of the embedding $\mathbf{z_{t_0}}$ with the original\\ data distribution.
\end{tabular}\\\hline

$\mathcal{L}_{adv}$& \begin{tabular}[c]{p{200pt}}
Adversarial loss term to regularize both generator and discriminator to map the produced graphs by the previous generator at $t_{i-1}$ onto the ground-truth ones at timepoint $t_{i}$.
\end{tabular}\\\hline

$\mathcal{L}_{1}$& \begin{tabular}[c]{p{200pt}}
Mean absolute error computed between two consecutive predicted graphs.
\end{tabular}\\\hline

$\mathcal{L}_{KL}$& \begin{tabular}[c]{p{200pt}}
Kullback-Leibler (KL) divergence to ensure the alignment of both distributions of the ground-truth and predicted brain graphs at timepoint $ti$.
\end{tabular}\\\hline

$\mathcal{L}_{1}(N)$& \begin{tabular}[c]{p{200pt}}
Mean absolute error that minimizes the distance between a normalized subject with respect to a CBT and its corresponding real brain graph.
\end{tabular}\\\hline

$\mathcal{L}_{sub}$& \begin{tabular}[c]{p{200pt}}
Subject integration loss ensuring the centeredness of a subject-based CBT (i.e., $\mathbf{C}_{s}$) computed in a specific cluster.
\end{tabular}\\\hline

$\mathcal{L}_{clt}$& \begin{tabular}[c]{p{200pt}}
Cluster integration loss ensuring the centeredness of a CBT estimated for the whole cluster.
\end{tabular}\\\hline

$\mathcal{L}_{SNL_{s}}$& \begin{tabular}[c]{p{200pt}}
Subject Normalization Loss (SNL) based on Forbinious distance introduced to evaluate the centerdness of the estimated CBT using a random subset of the training brain graphs.
\end{tabular}\\\hline

$\mathcal{L}_{ce}$ & \begin{tabular}[c]{p{200pt}}
Classification loss which cross entropy computed to regularize the GNN classifier of the brain graph.
\end{tabular}\\\hline 

$\mathcal{L}_{dist}^{(l)}$ & \begin{tabular}[c]{p{200pt}}
Maximum mean discrepancy or binary cross entropy loss terms used to assign distinguishable scores for the most relevant nodes and the irrelevant ones.
\end{tabular}\\\hline

$\mathcal{L}_{gdc}^{c}$, $\mathcal{L}_{TPK}^{(l)}$  & \begin{tabular}[c]{p{200pt}} Regularization group-level consistency loss term applied for the first pooling layer to ensure the identification of group of biomarkers reliable for disease classification such as assigning a score near to 1 for the most reliable brain region for a classification task and a score near to 0 for the non valuable ones.
\end{tabular}\\\hline

$\mathcal{L}_{reg1}$, $\mathcal{L}_{reg2}$, $\mathcal{L}_{reg}^{(l)}$ & \begin{tabular}[c]{p{200pt}} Regularization losses to optimize the pooling aggregation learning for the GNN and EGAT and BrainGNN classifiers, respectively.
\end{tabular}\\\hline

$I_{1}, I_{2}$ &  \begin{tabular}[c]{p{200pt}} Biomarker identification score so that a higher score means a distinguishable biomarker, and gradient sensitivity score to disentangle the relationship across features. 
\end{tabular}\\\hline

\begin{tabular}[c]{p{25pt}}$\mathcal{L}_{dis}$, $\mathcal{L}_{gen}$, $\mathcal{L}_{a}$ \end{tabular}& \begin{tabular}[c]{p{200pt}} Diagnosis and gender classification loss leveraging the cross-entropy between real and predicted labels and age prediction loss based on MSE metric.
\end{tabular}\\

\hline
\end{tabular}
\end{center}
\end{table}


\section{DISCUSSION AND OUTLOOK} 
Though significant advances have been made in the research of GNN models in the field of network neuroscience, there still exist many open problems that are worth exploring. In this section, we summarize our results from this review and critical challenges with possible future directions in this field.

\subsection{Toward clinical translation}
\hfill \break
\textbf{Strengths and limitations.} We have performed an extensive search to identify GNN-based methods applied to network neuroscience. We identified 30 papers from the 1st of January 2017 to the 31st of December 2020 within three different objectives: brain graph prediction, brain graphs integration, and disease classification. The list of these papers reviewed for our study can be found on our GitHub repository \footnote{https://github.com/basiralab}. About 36\% of these papers studied brain graph synthesis, with cross-domain, cross-resolution and cross-time synthesis being the most important application of GNNs. Morphological brain graph (MBG) is ranked as the most common brain modality explored in the studied literature. Specifically, the brain graph synthesis models were trained on brain graphs derived from the T1-weighted MRI. Considering that MBGs are derived from different brain views such as cortical thickness and sulcal depth, they are considered as multimodal brain graph synthesis works. We believe one of the reasons for the significant interest in predicting MBGs is its unique representation of brain connectivities in terms of similarity in morphology between brain regions that can reveal unseen patterns in structural and functional brain graphs \cite{Mahjoub2018,Soussia2018b}. Interestingly, all reviewed brain graph synthesis works are generic since they are designed to predict any type of brain graph from another given that they have the same size. Therefore, ongoing research should be adapted to non-isomorphic graphs derived from different modalities and brain parcellations. We further identified a single work with a goal of one-to-many graph synthesis and six one-to-one brain graph prediction works. Although important for alleviating the lack of connectomic data by predicting graphs from minimal resources, all reviewed brain graph synthesis works did not consider predicting graphs from multiple modalities that exist such as leveraging both structural and functional connectivities to predict a morphological one. This can be beneficial in the case of having a dataset with more than one modality (i.e., T1-w, fMRI, dMRI) so one might integrate them in a single framework to predict the missing graphs.  

\textbf{Reproducibility and explainability.} One of the key rationales behind brain graph prediction models is that synthesized graphs can boost the performance of existing models in early disease prediction and biomarker identification. We found 56\% of the GNN-based studies that fall into the group of disease classification. Within this category, only four papers addressed the interpretability of the implemented architectures. These works interpreted the biomarkers extracted from a single dataset and a single GNN classifier. This observation leads to a series of questions: if we train various GNN models for this task will they produce the same classification results and the same discovered biomarkers? Specifically, with the remarkable growth of brain connectomics datasets future works need to tackle the reproducibility of the learned GNNs across multiple datasets \cite{vandenHeuvel2019}. In other words, designing a framework that predicts which model consistently reproduces discriminative ROIs across different datasets will provide more generalizable clinical interpretations of brain disorders. Moreover, all of the existing works generated a the brain connectome using a particular atlas such as Desikan-Killiany Atlas. While there are different ways for constructing brain connectomes none of the existing works investigated the influence of such cortical parcellation on the learning and predictive performances of GNN models --with the exception of the cross-resolution works which are fundamentally designed to predict a brain connectome generated from a low-resolution parcellation atlas onto a high-resolution one. Generally, brain graph constructions methods use different parcellations to precisely delineate neural populations. Here are some examples: (1)  voxel-node parcellation where an MRI voxel is treated as a brain region in the brain graph \cite{van2008small}, (2) anatomical parcellation which is based on sulcal and gyral boundries to define the brain regions \cite{desikan2006automated}, (3) random parcellation which results in brain regions with approximately equal size \cite{fornito2010network}. The variation of the cortical parcellation methods leads to a difference in the constructed brain graphs where a set of nodes in one connectome might be considered as a single node in another connectome. Therefore, studying the reproducibility of existing GNN-based frameworks is important for examining the consistency of the predictive results regardless the parcellation scheme used to create the brain graphs. Further 6\% (2 papers) of these works are related to CBT estimation. CBT represents a tool for not only disentangling the connectivity patterns of healthy and disordered populations but also for augmenting brain graphs through localized perturbations of the CBT. Existing data augmentation operations are manually designed operations and are not able to cover the whole variability of the data. Therefore, leveraging the brain template to generate new connectomes will further pave the way for synthesizing brain graphs when minimal resources exist. However, explaining why these models estimate a well-representative template for a given population still untapped. This explanation can help identify connectivity patterns of the input brain graphs relevant to the model's decisions in the integration task \cite{explain1}. Therefore reproducibility \cite{reproduce1, reproduce2} and explainability \cite{explain1, gnnexplainer, graphlime} on brain graphs is an important direction to investigate in network neuroscience.

\textbf{GNN selection and learning improvement.} A key challenge hampering the field is that it is not always known a priori which GNN model should be used for a specific task. Here, we identified 50\% of the frameworks are based on GCN while the other 50\% are based on various graph convolution operations (e.g., ECC, NNConv, GATConv). While GCN was designed to learn from graphs having node features, it was leveraged for both population-based and graph-based models. Consequently, a key question arises: how to choose a GNN model to learn from brain graphs? Additionally, 53\% of the GCN works are GAN-based frameworks trained with different loss functions (\textbf {Table \ref{prediction}, \textbf {Table \ref{classification}}}). We found that some works added a topological loss term to the overall function to ensure the conservation of hubness and modularity properties of brain graphs. They mainly refer to centrality measures from the graph theory field such as closeness and betweenness and seemed to perform better than the basic GNN models that do not include such loss terms. Several important and unresolved questions about graph learning are raised: Are propagation rules of existing GNNs not able to learn the local node properties of the graph? Is there any evidence of choosing a specific centrality metric? Future work might include efforts in providing more effective GNN without the addition of topological loss terms to learn the deep graph structure.

\textbf{Few-shot learning.} And another key question looms on the horizon: still these models accurate when trained on a small dataset? Conventionally, the reviewed GNN models were learned on a large number of brain graphs which hinders their application to a few-shot learning (FSL) setting where the goal is to seek a better model generalization on problems with very limited data. While several image-based works have demonstrated the feasibility of learning given few samples \cite{FSL1, FSL2, FSL3}, FSL remains overlooked in brain graph-based works \cite{FSL}. Therefore, a promising direction is to consider the use of FSL in network neuroscience works. Although existing GNN models are designed to learn from a geometric data type that is brain graph, one might ask whether they are applied on other types such as cortical surfaces \cite{surface1}.  Would it be possible to generate cortical surfaces using the reviewed graph prediction works with the same time and accuracy performances? One can make an intuitively reasonable claim that brain manifold can be considered as graphs thus all mentioned models can learn on such data representation, but one needs to study the scalability and robustness of existing GNN models. However, existing studies \cite{mesh1} leveraged CNN for learning on 3D mesh which might not learn the local neighborhood. Specifically, meshes have a similar structure to graphs as it includes nodes and edges except that it does not denote the relationship between brain regions. For example, \cite{mesh2} proposed MeshCNN architecture for classification and segmentation of 3D meshes where mesh convolution and mesh pooling operations were introduced. Therefore, such observation will probably be a key focus of future GNN work that might be a combination study of graphs and manifolds that will further boost the diagnosis of various diseases.

\subsection{Outlook}
Perhaps the holy grail of brain graphs is to increase its clinical utility for various brain disorders. Recent progress in graph neural networks has shed new light on early diagnosis by synthesizing brain graphs across the different axes (i.e., domain, time, and resolution) from minimal connectomic data. Graph neural network models have achieved state-of-the-art performance across different medical applications; however, there is still room for improvement. First, as witnessed in several medical imaging and connectomic applications, in which breakthrough improvements were challenging due to the scarcity of the datasets, multimodal brain graph synthesis framework where three axis will be involved jointly would lead to improved performance of diagnosis-based models. Second, it will be important to achieve a greater understanding of the GNN models; in particular, we need to be able to explain how such models generate the desired results and more precisely how to select the GNN model that is more reproducible when learned on multiple datasets \cite{explain1,gnnexplainer}. It will also be important to investigate further the link between graph theory and graph representation learning fields for the sake of improving the learning of the models \cite{gtheory1, gtheory2}. Furthermore, it is necessary to develop GNNs that are trained with a frugal setting where a few brain graphs will be used in the learning step of the model as in  \cite{guvercin2021one,ozen2021flat,pala2021template}. {Finally, there are not currently available studies that motivate and explain the advantages and disadvantages of GNN in comparison to the other ML-based techniques such as \cite{ye2015intrinsic,cacciola2017coalescent}. Future studies might investigate the reason to adopt GNN in contrast to these other popular ML-based methods.}

\section*{Acknowledgment}
This project has been funded by the 2232 International Fellowship for Outstanding Researchers Program of TUBITAK (Project No:118C288). However, all scientific contributions made in this project are owned and approved solely by the authors. 

\bibliography{biblio}
\begin{IEEEbiography}[{\includegraphics[width=1in,height=1.25in,clip,keepaspectratio]{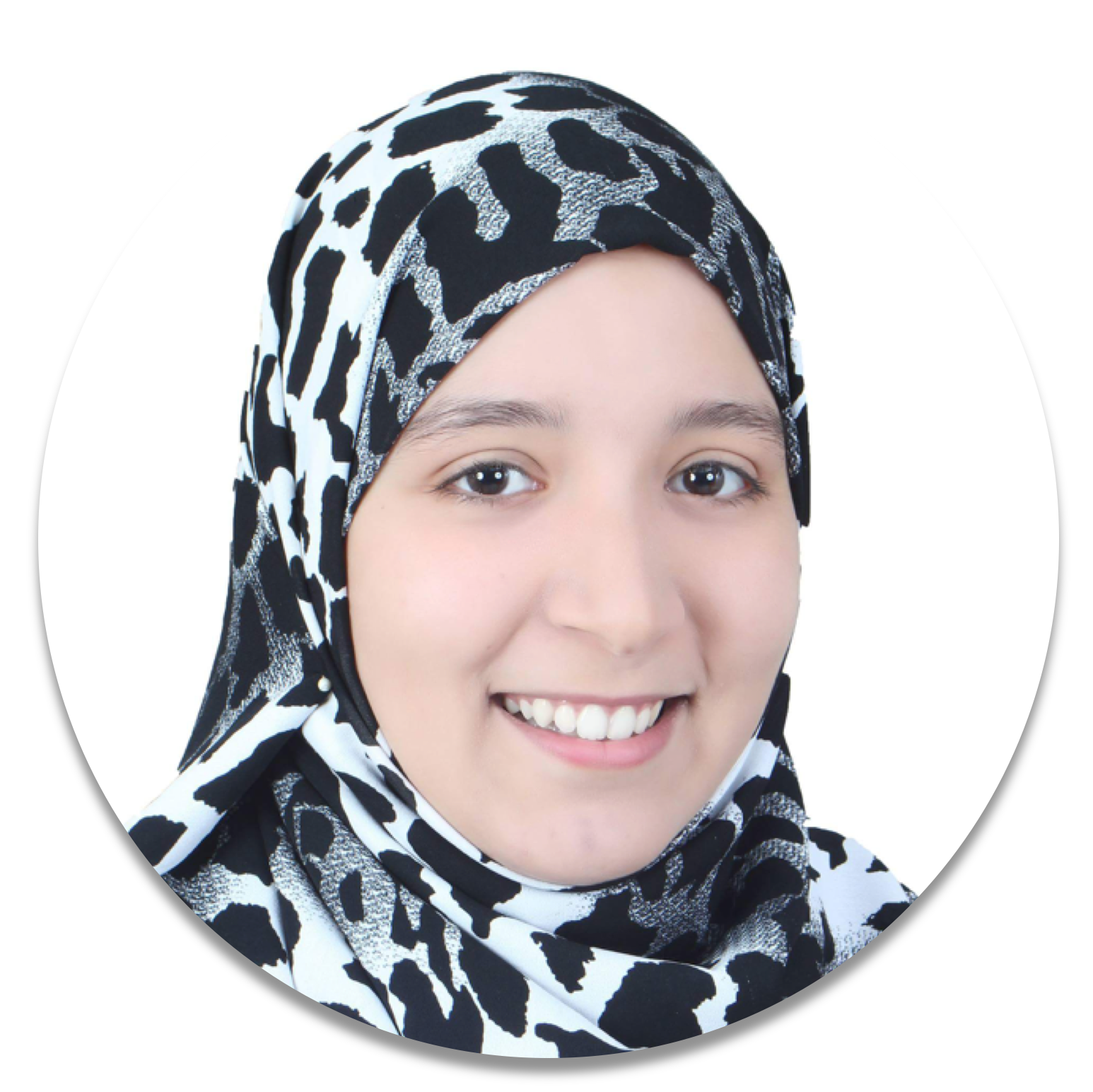}}]{Alaa Bessadok}
	PhD student at the Higher Institute of Computer Science and Communication Techniques (ISITCom), University of Sousse of Tunisia. She is co-affiliated with LATIS and BASIRA laboratories. She obtained her master and bachelor degrees from the Higher Institute of Management of Gabes from the University of Gabes of Tunisia in 2018 and 2015, respectively. In two successive years, she has won the MICCAI 2019 (Shenzhen, China) and 2020 Student Travel Award for having the first paper accepted from the continent of Africa in the history of MICCAI and for attending the virtual conference (Lima, Peru). She was a PC member of MICCAI (2020-2021) and MIDL 2021 conferences as well as MICCAI workshops including PRIME, FAIR and the NeurIPS workshop ``Medical Imaging meets NeurIPS''.
\end{IEEEbiography}

\begin{IEEEbiography}[{\includegraphics[width=1in,height=1.25in,clip,keepaspectratio]{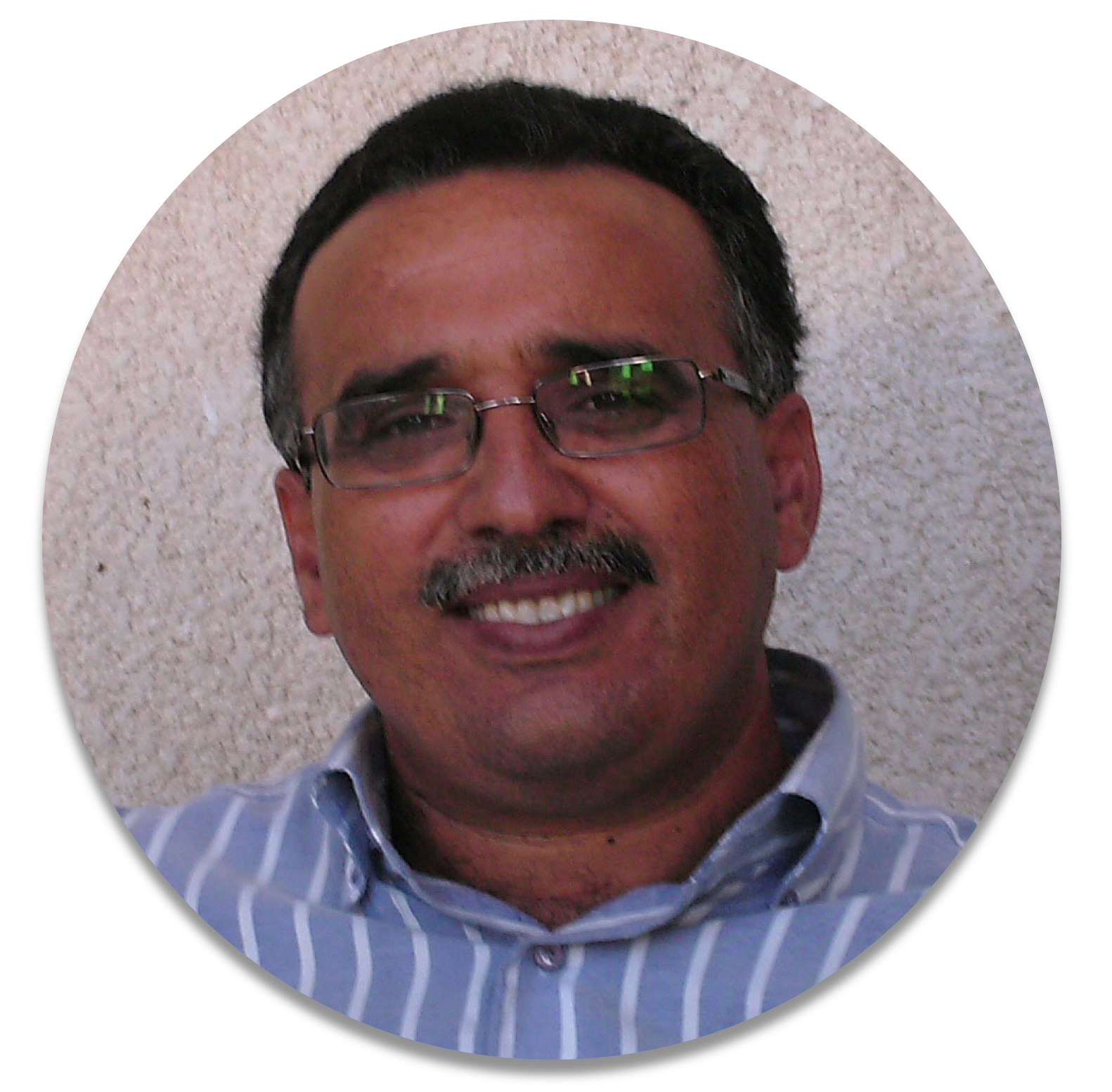}}]{Mohamed Ali Mahjoub}
	He received his Bsc and MSc in computer science from the National School of Computer Science of Tunis, Tunisia and National School of Engineers of Tunis, in 1990 and 1993 respectively. He obtained his Ph.D and HDR degrees in electrical  engineering and signal processing from the National School of Engineers of Tunis, Tunisia, in 1999 and 2013 respectively. From 1990 to 1999, he was a computer science engineer at the National School of Engineers of Tunis. In September 1999, he joined the preparatory institute of engineering of Monastir as an assistant professor in computer science. In December 2013, he joined the Electronics Engineering Department of the National School of Engineers of Sousse (ENISo) as a professor. He is a founding member of the Laboratory of Advanced Technology and Intelligent Systems (LATIS) at ENISo. His research interests include machine learning, probabilistic graphical models, computer vision, and pattern recognition.
\end{IEEEbiography}

\begin{IEEEbiography}[{\includegraphics[width=1in,height=1.25in,clip,keepaspectratio]{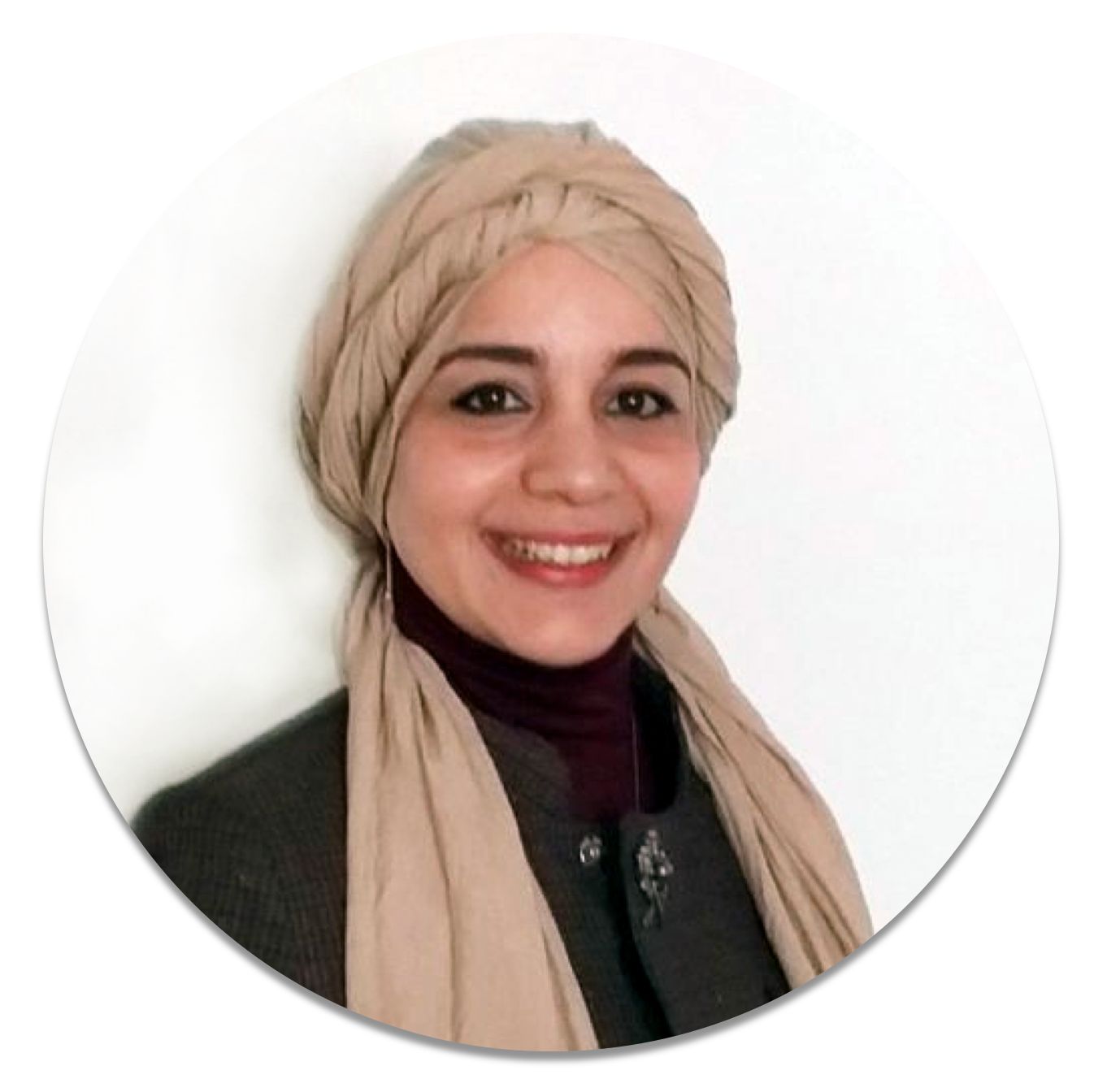}}]{Islem Rekik}
	Assistant Professor and Director of the Brain And SIgnal Research and Analysis (BASIRA) laboratory (\url{http://basira-lab.com/}), at Istanbul Technical University. She is the former president of the Women in MICCAI (\href{http://www.miccai.org/about-miccai/women-in-miccai/}{WiM}) and the co-founder of the international \href{http://www.miccai.org/about-miccai/rise-miccai/}{RISE Network} to Reinforce Inclusiveness \& diverSity and Empower minority researchers in Low-Middle Income Countries (LMIC). She completed her PhD at the University of Edinburgh in 2014, UK, and enjoyed two years of post-doctoral fellowship at IDEA lab, University of North Carolina from 2014-16. In 2019, she was awarded the 3-year prestigious TUBITAK 2232 for Outstanding Experienced Researchers Fellowship and in 2020 she became a Marie Sklodowska-Curie Fellow under the European Horizons 2020 program. She served as a PC member of MICCAI, IPMI, ISBI, area chair of MICCAI (2019-2021), and MIDL (2019). She is a member of the organizing committee of MICCAI 2023 (Vancouver) and 2024 (Marrakesh), the proposer/co-chair (2018-2021) of the MICCAI Workshop on Predictive Intelligence in MEdicine (\href{https://basira-lab.com/prime-miccai-2022/}{PRIME}), a co-chair of MICCAI workshops, including \href{http://miccai.brainconnectivity.net/}{CNI}, \href{https://sites.google.com/view/fair-workshop-2021/home}{FAIR}, and \href{https://sites.google.com/view/mlmi2022/}{MLMI} as well as NeurIPS workshops including \href{https://sites.google.com/view/med-neurips-2021}{``Medical Imaging meets NeurIPS''} 2021 and 2022.  She is the author of more than 130 peer-reviewed scientific papers, including more than 10 awarded papers (e.g., 5 at MICCAI and 1 at ATSIP).
\end{IEEEbiography}

\bibliographystyle{IEEEtran}
\end{document}